\documentclass[11pt]{article}
\usepackage[letterpaper,margin=1in,footskip=24pt]{geometry}
\usepackage[utf8]{inputenc}
\usepackage[T1]{fontenc}
\usepackage[english]{babel}
\usepackage{pifont}
\usepackage[numbers,super,sort,compress]{natbib}
\usepackage[colorlinks]{hyperref}
\usepackage{mathtools}
\usepackage[capitalise,noabbrev,nameinlink]{cleveref}
\usepackage[hypcap=false]{caption}
\usepackage[hang,flushmargin,symbol]{footmisc}
\usepackage{booktabs}
\usepackage{array}
\usepackage{fancyhdr}
\usepackage{xcolor}
\usepackage{xparse}
\usepackage{xpatch}
\usepackage{tabularray}
\usepackage{environ}
\usepackage{changepage}
\usepackage{lineno}
\usepackage{subcaption}
\usepackage{graphicx}
\usepackage{amsfonts}
\usepackage{nicefrac}
\usepackage{enumitem}
\usepackage{titlesec}
\usepackage{xspace}
\usepackage{wrapfig}
\usepackage[noend]{algpseudocode}
\usepackage{algorithm}

\RequirePackage{XCharter}
\RequirePackage[xcharter,bigdelims,vvarbb]{newtxmath}
\RequirePackage[scaled=1.1]{zlmtt}

\newcommand{\method}{Dreamer~4\xspace}

\fancypagestyle{plain}{%
\fancyhead[L]{\includegraphics[width=90pt]{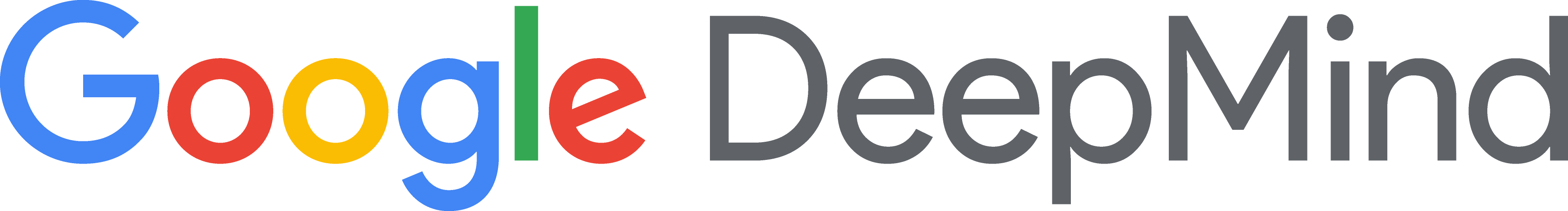}}
\fancyhead[C]{}
\fancyhead[R]{}
\fancyfoot[L]{\small%
* Equal contribution.\enskip
Google DeepMind, San Francisco, USA.\enskip
Website: \href{https://danijar.com/dreamer4}{danijar.com/dreamer4}
}
\fancyfoot[C]{}
\fancyfoot[R]{\small\thepage}

}

\fancypagestyle{fancy}{%
\fancyhead[L]{}
\fancyhead[C]{}
\fancyhead[R]{}
\fancyfoot[L]{}
\fancyfoot[C]{}
\fancyfoot[R]{\small\thepage}

}

\makeatletter
\renewcommand{\maketitle}{%
\thispagestyle{plain}%
\bgroup%
\begin{center}%
\vspace*{0pt}%
{\bfseries\fontsize{20pt}{20pt}\selectfont\@title\par}
\vspace*{12pt}%
{\normalsize\@author\par}
\vspace*{10pt}%
\end{center}%
\egroup}
\makeatother

\titleformat{\section}{\normalfont\Large\bfseries}{\thesection.}{.5em}{}
\titleformat{\subsection}{\normalfont\large\bfseries}{\thesubsection.}{.5em}{}

\definecolor{linkcolor}{rgb}{0.0,0.15,0.7}
\hypersetup{colorlinks=true,allcolors=linkcolor}
\setlist[itemize]{leftmargin=1.2em,labelsep=.7em,itemsep=0ex,topsep=0ex}
\renewcommand{\paragraph}[1]{\textbf{#1}\hspace{3ex}\removeParAfter}

\UseTblrLibrary{booktabs}
\NewTblrColumnType{L}[1]{Q[l,m,wd=#1]}
\NewTblrColumnType{C}[1]{Q[c,m,wd=#1]}
\NewTblrColumnType{R}[1]{Q[r,m,wd=#1]}
\renewcommand{\o}{\hphantom{0}}
\NewEnviron{mytabular}[1]{%
\setlength\heavyrulewidth{1.2pt}
\setlength\lightrulewidth{.5pt}
\setlength\aboverulesep{.4ex}%
\setlength\belowrulesep{.8ex}%
\begin{booktabs}[expand=\BODY]{#1}
\BODY
\end{booktabs}
}

\crefname{equation}{}{}
\crefname{algocf}{Algorithm}{Algorithms}
\Crefname{algocf}{Algorithm}{Algorithms}

\makeatletter
\newcommand{\algmargin}{\the\ALG@thistlm}
\makeatother
\algrenewcommand{\alglinenumber}[1]{$\bullet$}
\algdef{SxnE}[REPEATPLAIN]{RepeatPlain}{End}[0]{\algorithmicrepeat}{}

\algnewcommand{\ParState}[1]{\State\parbox[t]{\dimexpr\linewidth-\algmargin}{\strut#1\strut}}
\algnewcommand{\ParStateIndent}[1]{\State\quad\parbox[t]{\dimexpr\linewidth-\algmargin}{\strut#1\strut}}

\makeatletter
\DeclareDocumentCommand\todo{g}{%
\def\@message{\IfNoValueTF{#1}{TODO}{TODO: #1}}
\textbf{\textcolor[HTML]{FF8811}{\@message}}
\@latex@warning{\@message}{}{}}
\makeatother

\makeatletter
\newcommand{\removeParBefore}{\ifvmode\vspace*{-\baselineskip}\setlength{\parskip}{0ex}\fi}
\newcommand{\removeParAfter}{\@ifnextchar\par\@gobble\relax}
\newcommand{\eq}{\begingroup\removeParBefore\endlinechar=32 \eqinner}
\newcommand{\eqinner}[2][aligned]{\endlinechar=32%
\begin{gather}\begin{#1}#2\end{#1}\end{gather}\endgroup\removeParAfter}
\makeatother

\DeclareDocumentCommand{\p}{ D<>{p} D<>{} r() }{%
\def\content{#3}\patchcmd{\content}{|}{\;#2\vert\;}{}{}
\ensuremath{#1 #2(\content #2)}}

\DeclareDocumentCommand{\E}{ D<>{E} E{_}{{}} D<>{\big} r[] }{%
\def\content{#4}\patchcmd{\content}{|}{\;#3\vert\;}
{}\ensuremath{\operatorname{#1}_{#2}#3[\content #3]}}

\renewcommand{\o}{\hphantom{0}}
\newcommand{\sg}{\operatorname{sg}}

\newcommand{\dhalf}{\textstyle\frac{d}{2}}
\newcommand{\sig}{\tau}  
\newcommand{\dis}{d}

\title{\vspace*{-1.7ex}Training Agents Inside of Scalable World Models\vspace*{-.2ex}}

\date{}

\author{%
Danijar Hafner*\quad
Wilson Yan*\quad
Timothy Lillicrap
}

\begin{document}

\maketitle
\pagestyle{fancy}

\vspace*{-2ex}
{\ignorespaces\bfseries%
World models learn general knowledge from videos and simulate experience for training behaviors in imagination, offering a path towards intelligent agents.
However, previous world models have been unable to accurately predict object interactions in complex environments.
We introduce \method, a scalable agent that learns to solve control tasks by reinforcement learning inside of a fast and accurate world model.
In the complex video game Minecraft, the world model accurately predicts object interactions and game mechanics, outperforming previous world models by a large margin.
The world model achieves real-time interactive inference on a single GPU through a shortcut forcing objective and an efficient transformer architecture.
Moreover, the world model learns general action conditioning from only a small amount of data, allowing it to extract the majority of its knowledge from diverse unlabeled videos.
We propose the challenge of obtaining diamonds in Minecraft from only offline data, aligning with practical applications such as robotics where learning from environment interaction can be unsafe and slow.
This task requires choosing sequences of over 20,000 mouse and keyboard actions from raw pixels.
By learning behaviors in imagination, \method is the first agent to obtain diamonds in Minecraft purely from offline data, without environment interaction.
Our work provides a scalable recipe for imagination training, marking a step towards intelligent agents.
}

\vfill
\vfill
\begin{figure}[h!]
\centering
\includegraphics[width=\linewidth]{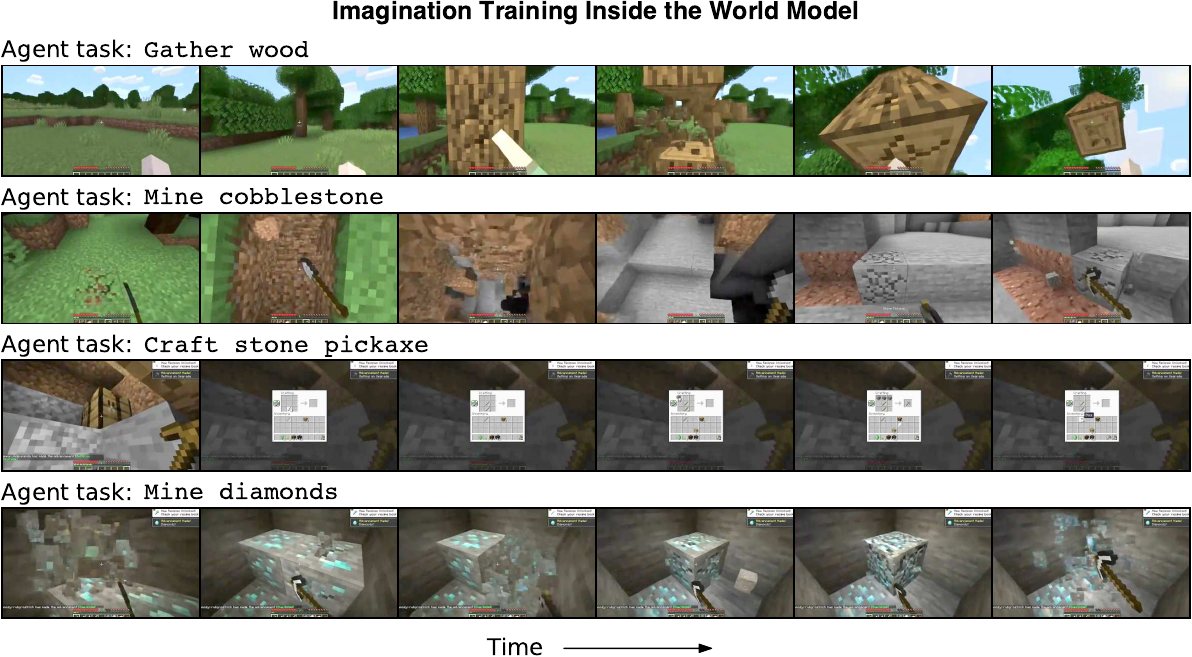}
\vspace*{-3ex}
\caption{\method learns to solve complex control tasks by reinforcement learning inside of its world model.
We decode the imagined training sequences for visualization, showing that the world model has learned to simulate a wide range of game mechanics from low-level mouse and keyboard actions, including breaking blocks, using tools, and interacting with crafting tables.
}
\label{fig:imag}
\vspace*{-1.5ex}
\end{figure}

\vfill
\clearpage

\section{Introduction}

To solve complex tasks in embodied environments, intelligent agents need to deeply understand the world and choose successful actions.
World models offer a promising approach towards this goal by learning to predict the future outcomes of potential actions from the perspective of an agent, such as a robot or a video game player.
This way, world models equip agents with a deep understanding of the world and the ability to choose actions by planning or reinforcement learning in imagination.
Moreover, world models can in principle learn from fixed datasets, allowing to train agents purely in imagination without the need for online interaction.
Optimizing behaviors offline is valuable for many practical applications, such as robots in the physical world, where online interaction with a partially trained agent is often unsafe.

World model agents, such as Dreamer~3, are among the best-performing and most robust reinforcement learning algorithms for games and robotics to date \citep{dreamerv3,wu2023daydreamer,hansen2023tdtmpc2,alonso2024diffusion,schrittwieser2019muzero,hessel2021muesli}.
While these models are fast and accurate for their narrow environments, their architecture lacks the ability to fit complex real world distributions.
Controllable video models, such as Genie~3, have been trained on diverse real video and games and have accomplished diverse scene generation and simple interactions \citep{genie3,tu2025playerone,he2025matrix,sun2025virtual,bai2025whole,team2025yan}.
These models are based on scalable architectures, such as diffusion transformers\citep{peebles2023dit,diffusionforcing}.
However, they still struggle to learn the precise physics of object interactions and game mechanics, limiting their usefulness for training successful agents.
Moreover, they often require many GPUs to simulate a single scene in real time, further reducing their practicality for imagination training.

We introduce \method, a scalable agent that solves control tasks by imagination training inside of a fast and accurate world model.
\method is the first agent to obtain diamonds in the challenging video game Minecraft purely from a standard offline dataset, without environment interaction.
\method leverages a novel shortcut forcing objective and an efficient transformer architecture to accurately learn complex object interactions while enabling real-time human interaction and efficient imagination training.
We show that the world model accurately predicts a wide range of semantic interactions in Minecraft, outperforming previous world models by a large margin.
Moreover, \method can be trained on large amounts of unlabeled videos and requires only a small amount of videos paired with actions.
This opens up the possibility of learning general world knowledge from diverse web videos in the future, for which action labels are not available.

Our contributions are summarized as follows:

\begin{itemize}
\item We introduce \method, a scalable agent that learns to solve challenging control tasks by imagination training inside of a world model.
\item \method is the first agent to collect diamonds in Minecraft from only offline data, substantially improving over OpenAI's VPT offline agent\citep{vpt} despite using 100$\times$ less data.
\item We introduce a high-capacity world model that achieves real-time inference on a single GPU through a shortcut forcing objective and an efficient transformer architecture.
\item We show that the world model accurately predicts a wide range of object interactions and game mechanics in Minecraft, substantially outperforming previous world models.
\item We show that the world model can learn from unlabeled videos and requires only a small amount of aligned data to learn action conditioning with strong generalization.
\item An extensive ablation study measures the improvements of the objective and architecture.
\end{itemize}

\section{Background}
\label{sec:background}

\paragraph{Flow matching}

Our world model is based on the paradigm of diffusion models~\citep{sohl2015deep,ddpm}, where the network $f_\theta$ is trained to restore the a data point $x_1$ given a corrupted version $x_\sig$.
The signal level $\sig \in [0, 1]$ determines the mixture of noise and data and is randomized during training, where $\sig=0$ corresponds to pure noise and $\sig=1$ means clean data.
We build on the flow matching formulation~\citep{flowmatching,rectifiedflow} because of its simplicity, where the network predicts the velocity vector $v = x_1 - x_0$ that points towards the clean data:

\eq{
\begin{gathered}
x_\sig = (1 - \sig)\,x_0 + \sig\,x_1 \qquad
x_0 \sim \operatorname{N}(0, \mathbb{I}) \qquad
x_1 \sim \mathcal{D} \qquad
\sig \sim \p(\sig) \\[1.5ex]
\mathcal{L}(\theta) = \| f_\theta(x_\sig,\sig) - (x_1 - x_0) \|^2
\end{gathered}
}

The signal level is typically sampled from a uniform distribution or a logit-normal distribution \citep{sd3}.
At inference time, the sampling process starts with a pure noise vector $x_0$ and iteratively transforms it into a clean data point $x_1$ over $K$ sampling steps with step size $d = 1/K$:

\eq{
x_{\sig+\dis} = x_\sig + f_\theta(x_\sig,\sig)\,\dis \qquad
x_0 \sim \operatorname{N}(0, \mathbb{I})
}

\paragraph{Shortcut models}

Shortcut models \citep{shortcut} condition the neural network not only on the signal level $\sig$ but also on the requested step size $\dis$.
This allows them to choose the step size at inference time and generate data points using only a few sampling steps and forward passes of the neural network.
For the finest step size $\dis_\mathrm{min}$, shortcut models are trained using the flow matching loss.
For larger step sizes $\dis_\mathrm{min}<\dis \leq 1$, shortcut models are trained using a bootstrap loss that distills two smaller steps, where $\operatorname{sg}(\cdot)$ stops the gradient:

\eq{
\begin{gathered}

x_0 \sim \operatorname{N}(0, \mathbf{I}) \qquad
x_1 \sim \mathcal{D} \qquad
\sig,\dis \sim \p(\sig,\dis) \\[1.2ex]

b' = f_\theta(x_\sig,\sig,\dis/2) \qquad
b'' = f_\theta(x',\sig+\dis/2,\dis/2) \qquad
x' = x_\sig + b'\,\dis/2 \\

\mathcal{L}(\theta) = \| f_\theta(x_\sig,\sig,\dis) - v_\mathrm{target} \|^2
\qquad
v_\mathrm{target} = \begin{cases}
x_1 - x_0 &\text{if } \dis=\dis_\mathrm{min}  \\
\sg(b' + b'')/2 &\text{else}
\end{cases}

\end{gathered}
}

The step size is sampled uniformly as a power of two, based on the maximum number of sampling steps $K_\mathrm{max}$, which defines the finest step size $\dis_\mathrm{min}=1/K_\mathrm{max}$. The signal level is sampled uniformly over the grid that is reached by the current step size:

\eq{
\dis \sim 1/\operatorname{U}(\{1, 2, 4, 8, \dots, K_{\mathrm{max}}\}) \qquad
\sig \sim \operatorname{U}(\{0, 1/\dis, \dots, 1 - 1/\dis\})
}

At inference time, one can condition the model on a step size $\dis=1/K$ to target $K$ sampling steps, without suffering from discretization error because the model has learned to predict the end point of each step. In practice, shortcut models generate high-quality samples with 2 or 4 sampling steps, compared to 64 or more steps for typical diffusion models.

\paragraph{Diffusion forcing}

For sequential data, diffusion forcing \citep{diffusionforcing} assigns a different signal level to each time step of the data sequence, producing a corrupted sequence.
This allows applying loss terms to all time steps in the sequence, where each time step serves both as denoising task and as history context for later time steps.
At inference time, diffusion forcing supports flexible noise patterns, such as generating the next frame given clean or lightly noised history.

\pagebreak
\section{World Model Agent}

\begin{wrapfigure}[21]{R}{0.44\textwidth}
\vspace*{-5ex}\hfill%
\begin{minipage}{0.43\textwidth}
\begin{algorithm}[H]
\caption{\enskip\method}
\label{alg:agent}
\begin{hyphenrules}{nohyphenation}

\begin{algorithmic}[1]
\setlength{\itemsep}{.5ex}

\vspace{1ex}
\Statex \hspace*{-2.3ex}\textbf{Phase 1:} World Model Pretraining
\State Train tokenizer on videos using \cref{eq:tok}.
\State Train world model on tokenized videos and optionally actions using \cref{eq:dyn}.

\vspace{1ex}
\Statex \hspace*{-2.3ex}\textbf{Phase 2:} Agent Finetuning
\State Finetune world model with task inputs for policy and reward heads using \cref{eq:dyn} and \cref{eq:bcrm}.

\vspace{1ex}
\Statex \hspace*{-2.3ex}\textbf{Phase 3:} Imagination Training
\State Optimize policy head using \cref{eq:pol} and value head using \cref{eq:val} on trajectories generated by the world model and the policy head.
\vspace{1ex}

\end{algorithmic}
\end{hyphenrules}
\end{algorithm}
\end{minipage}
\end{wrapfigure}
\begin{figure}[t!]
\vspace*{-2\baselineskip}
\centering
\begin{subfigure}[t]{0.45\textwidth}
\centering
\includegraphics[height=2.4in]{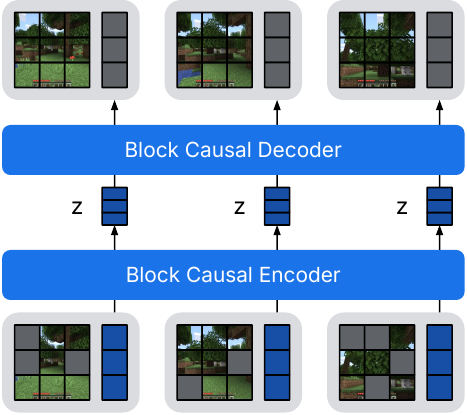}
\caption{Causal Tokenizer}
\end{subfigure}%
\hfill%
\begin{subfigure}[t]{0.45\textwidth}
\centering
\includegraphics[height=2.5in]{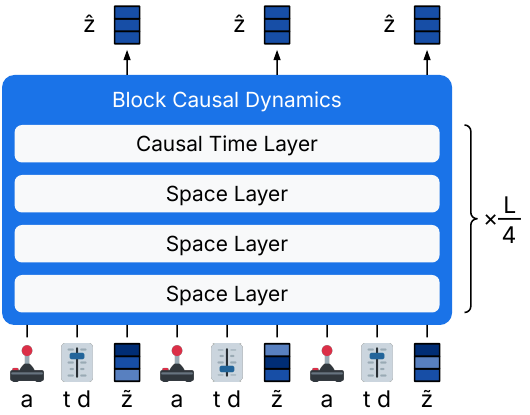}
\caption{Interactive Dynamics}
\end{subfigure}
\caption{World model design.
\method consists of a causal tokenizer and an interactive dynamics model, which both use the same block-causal transformer architecture.
The tokenizer encodes partially masked image patches and latent tokens, squeezes the latents through a low-dimensional projection with tanh activation, and decodes the patches.
It uses causal attention to achieve temporal compression while allowing frames to be decoded one by one.
The dynamics model operates on the interleaved sequence of actions, shortcut noise levels and step sizes, and tokenizer representations.
It denoises representations via a shortcut forcing objective.
After pretraining, the world model is finetuned into an agent by inserting task tokens into the dynamics transformer and predicting actions, rewards, and values from them.
}
\label{fig:model}
\end{figure}

We present \method, a scalable agent that learns to solve complex control tasks by reinforcement learning inside of a fast and accurate world model.
The agent consists of a tokenizer and a dynamics model, as shown in \cref{fig:model}.
The tokenizer compresses video frames into continuous representations and the dynamics model predicts the representations given interleaved actions, both using the same efficient transformer architecture.
The tokenizer is trained using masked autoencoding and the dynamics is trained using a shortcut forcing objective to enable interactive generations with a small number of forward passes and prevent accumulating errors over time.
As outlined in \cref{alg:agent}, we first pretrain the tokenizer and world model on videos and actions, then finetune the policy and reward model into the world model by interleaving task embeddings, and finally post-train the policy through imagination training.
To train a single dynamics transformer with multiple modalities and output heads, we normalize all loss terms by running estimates of their root-mean-square (RMS).

\pagebreak
\subsection{Causal Tokenizer}
\label{sec:tokenizer}

The tokenizer compresses raw video into a sequence of continuous representations for the dynamics model to consume and generate.
It consists of an encoder and a decoder with a bottleneck in between.
Both components are causal in time, enabling temporal compression while maintaining the ability to decode frame by frame for interactive inference.

\paragraph{Architecture}

We use the efficient transformer architecture described later.
Each time step consists of patch tokens of the current image and learned latent tokens.
After applying the encoder, the representations are read out of the latent tokens using a linear projection to a smaller channel dimension followed by a \texttt{tanh} activation.
For the decoder, this representation is projected back up to the model dimension and concatenated with learned tokens to read out the patches.
To flexibly integrate multiple input modalities if available, the encoder allows the latent tokens to attend to all modalities, while each modality only attends within itself.
Correspondingly, each decoder modality attends within itself and to the latents, while the latents only attend within themselves.

\paragraph{Masked autoencoding}

We train the tokenizer using a straightforward reconstruction objective, consisting of mean squared error and LPIPS \citep{lpips} loss.
To simplify weighing the two loss terms, we employ loss normalization as explained later.

\eq{
\mathcal{L}(\theta) =
\mathcal{L}_{\mathrm{MSE}}(\theta) +
0.2\,\mathcal{L}_{\mathrm{LPIPS}}(\theta)
\label{eq:tok}
}

We drop out input patches to the encoder to improve its representations using masked autoencoding \citep{mae,chen2025maetok}.
The dropout probability is randomized across images as $p \sim U(0, 0.9)$. Patches of each image are replaced with a learned embedding with this probability, so that the tokenizer is sometimes trained on the $p=0$ case used during inference.
We found MAE training to improve the spatial consistency of videos generated by the dynamics model.

\subsection{Interactive Dynamics}
\label{sec:dynamics}

The dynamics model operates on the interleaved sequence of actions and representations produced by the frozen tokenizer.
It is trained using a shortcut forcing objective to enable fast interactive inference with $K=4$ forward passes per generated frame.

\paragraph{Architecture}

The dynamics model uses our efficient transformer architecture on interleaved blocks of observations and actions.
The representations are linearly projected into $S_\mathrm{z}$ spatial tokens and concatenated with $S_\mathrm{r}$ learned register tokens \citep{vitregister} and a single token for the shortcut signal level and step size.
Since the signal level and step size are discrete, we encode each with a discrete embedding lookup and concatenate their channels.
Actions can contain multiple components, such as mouse and keyboard.
We encode each action component separately into $S_\mathrm{a}$ tokens and sum the results together with a learned embedding.
Continuous actions components are linearly projected and categorical or binary components use an embedding lookup.
When training unlabeled videos, only the learned embedding is used.

\paragraph{Shortcut forcing}

For efficient training and inference, we train the dynamics model using a shortcut forcing objective, which builds on diffusion forcing \citep{diffusionforcing} and shortcut models \citep{shortcut}, reviewed in \cref{sec:background}.
We formulate the objective in data space to prevent accumulating errors caused by high-frequency network outputs and introduce a simple loss weight to focus the model capacity on the loss terms with the most learning signal.
The dynamics model takes the interleaved sequence of actions $a=\{a_t\}$, discrete signal levels $\sig=\{\sig_t\}$ and step sizes $\dis=\{\dis_t\}$, and corrupted representations $\tilde{z}=\{z_t^{(\tau_t)}\}$ as input and predicts the clean representations $z_1=\{z_t^1\}$.
Note that $t \in [1,T]$ is the sequence timestep while $\tau_t \in [0, 1]$ is the signal level at that step.

\eq{
\begin{gathered}
z_0 \sim \operatorname{N}(0, \mathbf{1}) \qquad
z_1 \sim \mathcal{D} \qquad
\sig, \dis \sim \p(\sig, \dis) \qquad
\sig, \dis \in [0, 1]^T \\[1ex]
\hat{z}_1 = f_\theta(\tilde{z},\sig,\dis,a) \qquad
\tilde{z} = (1-\sig)\,z_0 + \sig\,z_1
\end{gathered}
}

Shortcut models parameterize the network to predict velocities $v = x_1 - x_0$, called v-prediction \citep{kingma2023understanding}.
This approach excels when generating the output jointly as one block, such as for image or video generation models.
However, v-prediction trains the network to produce high-frequency outputs.
When iteratively generating long videos frame by frame, this can cause subtle errors that accumulate over time.
Instead, we found that parameterizing the network to predict clean representations, called x-prediction, enables high-quality rollouts of arbitrary length.
Computing the flow loss term in x-space is straightforward \citep{kingma2023understanding}. To compute the bootstrap loss term, we convert the network output into v-space and scale the resulting loss back into x-space\footnote{
The network output is converted as $\hat{v}_\sig = (\hat{x}_1 - x_\sig) / (1 - \sig)$. The MSE in x-space and v-space is related by $\|\hat{x}_1 - x_1\|^2_2 = (1-\sig)^2\|\hat{v}_\sig-v_\sig\|^2_2$, motivating a $(1-\sig)^2$ multiplier to bring the bootstrap loss into a range similar to the x-space flow loss.}:

\eq{
\begin{gathered}
\begin{aligned}
b' &= (f_\theta(\tilde{z},\sig,\dhalf,a) - z_\sig)/(1-\sig) \qquad
z' = \tilde{z} + b'\,\dhalf \\
b'' &= (f_\theta(z',\sig+\dhalf,\dhalf,a) - z')/(1-(\sig+\dhalf))
\end{aligned} \\
\mathcal{L}(\theta) = \begin{cases}
\|\hat{z}_1 - z_1\|_2^2 &\text{if } \dis=\dis_\mathrm{min} \\
(1-\sig)^2 \| (\hat{z}_1-\tilde{z})/(1-\sig) - \sg(b_1 + b_2)/2  \|_2^2 &\text{else} \\
\end{cases}
\end{gathered}
\label{eq:dyn}
}

Low signal levels contain less learning signal, because the flow matching term degenerates to predicting the dataset mean, while the bootstrap term is generally easier to optimize because it has deterministic targets compared to the noisy flow matching term.
To focus the model capacity on signal levels with the most learning signal, we propose a \texttt{ramp} loss weight that linearly increases with the signal level $\sig$, where $\sig=0$ corresponds to full noise and $\sig=1$ to clean data:

\eq{
w(\sig) = 0.9 \sig + 0.1
}

At inference time, the dynamics model supports different noise patterns.
We sample autoregressively in time and generate the representations of each frame using the shortcut model with $K=4$ sample steps with corresponding step size $\dis=1/4$.
We slightly corrupt the past inputs to the dynamics model to signal level $\sig_\mathrm{ctx}=0.1$ to make the model robust to small imperfections in its generations.

\subsection{Imagination Training}

To solve control tasks, we first adapt the pretrained world model to predict actions and rewards from the dataset conditioned on one of multiple tasks.
For this, we insert agent tokens as an additional modality into the world model transformer and interleave it with the image representations, actions, and register tokens.
The agent tokens receive task embeddings as input and we use them to predict the policy and reward model using MLP heads.
While the agent tokens attend to themselves and all other modalities, no other modalities can attend back to the agent tokens.
This is crucial for avoiding causal confusion of the world model---its future predictions can only be directly influenced by actions, not by the current task.
To improve beyond strategies displayed in the dataset, we then finetune the policy through imagination training by reinforcement learning on rollouts generated by the world model, using an additional value head.

\paragraph{Behavior cloning and reward model}

After pretraining the world model on action-conditioned video prediction, the second training phase involves learning a task-conditioned policy and reward model.
Given a dataset of videos $x=\{x_t\}$ that are encoded into representations $z=\{z_t\}$, actions $a=\{a_t\}$, tasks $q=\{q_t\}$, and scalar rewards $r=\{r_t\}$, we train the policy and reward heads on the task output embeddings $h_t$ using multi-token prediction (MTP) \citep{gloeckle2024mtp} of length $L=8$:

\eq{
\mathcal{L}(\theta) =
- \sum_{n=0}^{L} \p<\ln p_\theta>(a_{t+n} | h_t)
- \sum_{n=0}^{L} \p<\ln p_\theta>(r_{t+n} | h_t)
\label{eq:bcrm}
}

To preserve existing capabilities, we reuse the pretraining setting with this additional loss function, so the representations are noisy and we continue to apply the video prediction loss.
We parameterize the policy and reward heads using small MLPs with one output layer per MTP distance.
Following Dreamer~3, the reward head is parameterized as a symexp twohot output \citep{dreamerv3} to robustly learn stochastic rewards across varying orders of magnitude.
The policy head is parameterized as categorical or vectorized binary distribution, depending on the action space of the dataset.

\paragraph{Reinforcement learning}

To improve the policy beyond behaviors displayed in the dataset, we continue training it with reinforcement learning on imagined rollouts to maximize the learned reward model.
Unlike online reinforcement learning that requires interaction with the environment, our policy learns purely inside the world model, enabling it to improve offline.
We initialize a value head and a frozen copy of the policy head that serves as a behavioral prior.
We only update the policy and value heads and keep the transformer frozen.\footnote{Finetuning the full transformer provides small additional benefits at higher computational cost. For that, the dynamics, policy prior, and reward losses need to be applied during imagination training to preserve their functions.}
Imagined rollouts start from contexts of the dataset that was used during the earlier training phases.
Unlike previous generations of Dreamer, we start only one rollout from each context, prioritizing data diversity and reducing memory consumption.
The rollouts are generated by unrolling the transformer with itself, sampling representations $z=\{z_t\}$ from the flow head and actions $a=\{a_t\}$ from the policy head.
We annotate the resulting trajectories with rewards $r=\{r_t\}$ using the reward head and values $v=\{v_t\}$ using the value head.

The value head is trained to predict the discounted sum of future rewards, allowing the policy to maximize rewards beyond the imagination horizon.
It uses a symexp twohot output to robustly learn across different scales of values\citep{dreamerv3}.
We train the value head using temporal difference learning (TD-learning) \citep{sutton1988td} to predict $\lambda$-returns computed from the predicted rewards and values along the sequence, where $\gamma=0.997$ is a discount factor and $c_t$ indicates non-terminal states:

\eq{
\mathcal{L}(\theta) = -\sum_{t=1}^T \ln\p<p_\theta>(R^\lambda_t | s_t)
\qquad
R^\lambda_t = r_t + \gamma c_t \big((1-\lambda)v_t + \lambda R^\lambda_{t+1}\big)
\qquad
R^\lambda_T = v_T
\label{eq:val}
}

Unlike previous generations of Dreamer, the policy head learns using PMPO \citep{abdolmaleki2024pmpo}, a robust reinforcement learning objective that uses the sign of the advantages $A_t = R^\lambda_t-v_t$ and ignores their magnitude.
This property alleviates the need for normalizing returns or advantages and ensures equal focus on all tasks despite potentially differing return scales.
PMPO balances the focus on positive and negative feedback by separately averaging a simple maximum likelihood loss over the states with positive and negative advantages, respectively.
We assign all imagined states $s_i$ across batch and time dimensions to the positive set $\mathcal{D}^+=\{s_i \mid A_t \geq 0\}$ or the negative set $\mathcal{D}^-=\{s_i \mid A_t < 0\}$ and apply the following policy loss:

\eq{
\mathcal{L}(\theta) &=
\frac{1-\alpha}{|\mathcal{D}^-|} \sum_{i \in \mathcal{D}^-}  \ln \p<\pi_\theta>(a_i|s_i)
-\frac{\alpha}{|\mathcal{D}^+|} \sum_{i \in \mathcal{D}^+} \ln \p<\pi_\theta>(a_i|s_i)
+\frac{\beta}{N} \sum_{i=1}^N \operatorname{KL}[\p<\pi_\theta>(a_i|s_i)\,\|\,\pi_\mathrm{prior}] \\
\label{eq:pol}
}

We set $\alpha=0.5$ to balance the positive and negative sets equally and use a weaker scale of $\beta=0.3$ for the behavioral prior.
Unlike the original PMPO objective, we use a reverse direction for the prior KL to better constrain the policy to the space of reasonable behaviors.
We find the scaling between the three objective terms to be highly robust in practice as they are all measured in nats\citep{shannon1948infotheory}.

\subsection{Efficient Transformer}

Scaling up world models to diverse data distributions while maintaining fast inference requires an efficient high capacity architecture.
In this section, we introduce our efficient transformer architecture that is used for both the tokenizer and the dynamics model.
The architecture is a 2D transformer \citep{vaswani2017transformer} with time and space dimensions.
To support interactive generation, the attention is masked to be causal in time, so that all tokens within a time step can attend to each other and to the past.
We start from a standard transformer with pre-layer RMSNorm \citep{rmsnorm}, RoPE \citep{rope}, and SwiGLU \citep{swiglu}.
We employ QKNorm \citep{dehghani2023scaling} and attention logit soft capping \citep{bello2016neural,gemma2} to increase training stability.

\paragraph{Efficiency}

Inference speed of block-causal transformer architectures is limited both by the FLOPs of the MLPs and the memory bandwidth needed to access the KV cache of a long context to attend into.
We employ a sequence of improvements to speed up inference, some of which also accelerate training.
First, we break up the cost of dense attention over all video tokens by using separate space-only and time-only attention layers \citep{axial}.
Second, we find that only a relatively small number of temporal layers are needed and only use temporal attention once every 4 layers, in line with recent findings \citep{llama4}.
Third, we apply GQA \citep{gqa} to all attention layers in the dynamics, where multiple query heads attend to the same key-value heads to reduce the KV cache size further.

\paragraph{Sequence length}

Increasing spatial tokens directly improves visual quality, whereas increasing temporal tokens allows training longer context lengths for more temporally consistent generation.
To support efficient training, we alternate training on many short batches and occasional long batches, and finetune the model on only long batches afterwards.
Alternating batch lengths produces intermediate training metrics and generations that are more indicative of final model performance than training only on short batches.
The batch lengths need to be longer than the context length of the model to prevent the transformer from overfitting to always seeing a start frame at the beginning of its context, enabling length generalization to arbitrary generation lengths.

\pagebreak
\section{Experiments}

\begin{figure}[t!]
\centering
\vspace*{-3ex}
\includegraphics[width=\linewidth]{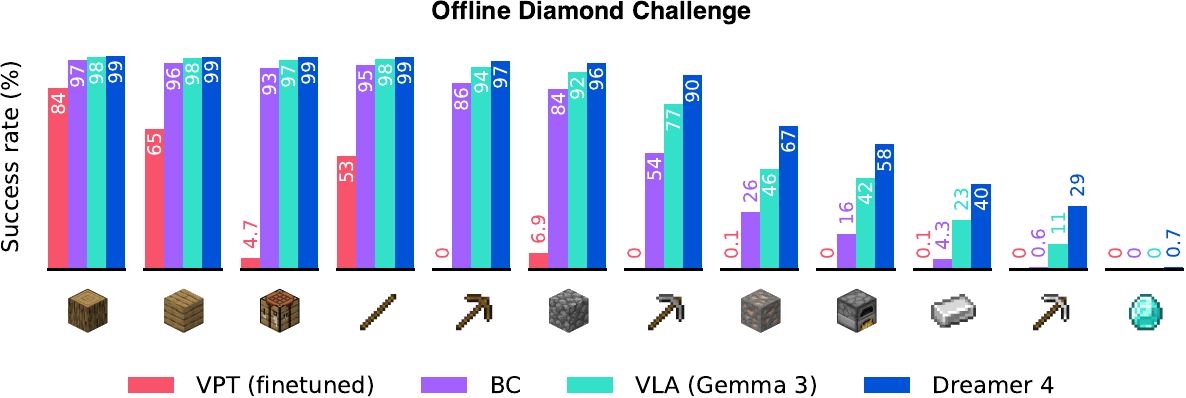}
\caption{
Agent performance in Minecraft without environment interaction.
All methods have access to the same contractor dataset\citep{vpt} with image inputs and low-level mouse and keyboard actions.
We report the success rates of important items obtained during 60-minute episodes that start in random worlds and from empty inventory, computed over 1000 episodes.
Leveraging imagination training, \method is the first agent to obtain diamonds in Minecraft purely from offline experience.
\method substantially outperforms OpenAI's VPT offline agent\citep{vpt} while using 100$\times$ less data.
It also outperforms our VLA agent \citep{kim2024openvla,intelligence2025pi05}, which leverages the general knowledge of the Gemma 3 vision-language model\citep{team2025gemma3}, nearly tripling its success rate for crafting iron pickaxes.
}
\label{fig:rl}
\end{figure}

We perform a wide range of experiments to evaluate and explore the capabilities of \method.
The majority of our experiments focus on Minecraft, a complex video game that features infinite open worlds including monsters and hundreds of items that can be mined or crafted, with raw pixel observations and low-level mouse and keyboard actions.
We primarily use the VPT dataset \citep{vpt} that contains 2541 hours of contractor gameplay with 360p video and mouse and keyboard actions at 20 FPS.
The experiments are designed to answer the following questions:

\begin{itemize}
\item Does \method learn to solve challenging control tasks purely by imagination training inside the world model, without online environment interaction? (\Cref{sec:control})
\item How well does \method learn to predict accurate object interactions and game mechanics in Minecraft compared to previous world models? (\Cref{sec:interact})
\item How much action data does \method need for learning action conditioning, and how far does the learned action grounding generalize? (\Cref{sec:actgen})
\item To what extent does each component of its objective and architecture contribute to the performance of \method? (\Cref{sec:modeldesign})
\end{itemize}

We train models with 2B parameters---400M for the tokenizer and 1.6B for the dynamics model---on 256 to 1024 TPU-v5p with batch size 1 per device and FSDP sharding \citep{fsdp,deepspeed}.
To improve generations without context, we treat 30\% of the videos in the batch as separate images, effectively training the dynamics model to generate start frames.
For Minecraft, we use 256 spatial tokens with 192 frames context length and 256 batch length.
For the real world datasets, we use 512 spatial tokens with 96 frames context length and 128 batch length.

\subsection{Offline Diamond Challenge}
\label{sec:control}

\begin{figure}[t!]
\centering
\vspace*{-2ex}
\includegraphics[width=\linewidth]{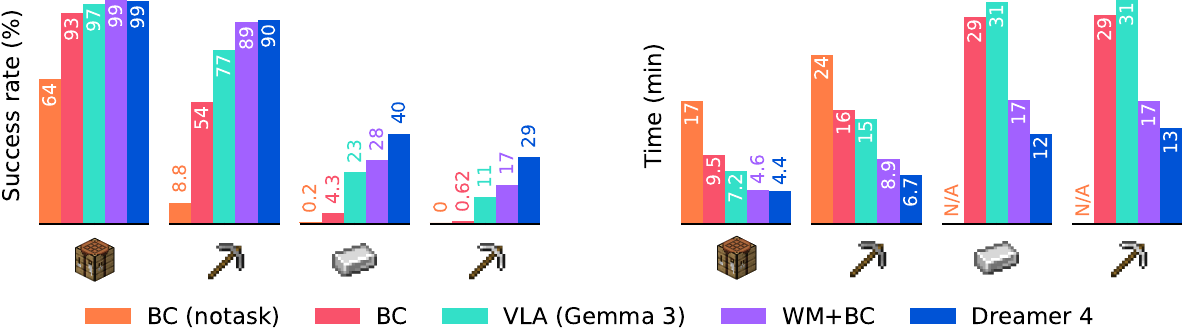}
\caption{
Agent ablations on the offline diamond challenge.
We report success rates and time needed to reach an item for four milestone items.
\method outperforms methods based on behavioral cloning in both metrics, demonstrating that imagination training improves both the robustness and efficiency of the policy.
Moreover, using the world model representations for behavioral cloning outperforms using Gemma 3 or training from scratch.
}
\label{fig:rlabl}
\end{figure}

We evaluate \method on the Minecraft diamond challenge, a long-horizon control task that requires solving several sub tasks, such as gathering materials and crafting tools in a complex procedurally generated 3D world from raw pixels and mouse and keyboard actions.
Human players with Minecraft experience take 20 minutes to collect a diamond on average, corresponding to sequences of 24,000 mouse and keyboard actions.

\paragraph{Offline setting}
While previous agents have achieved diamonds in Minecraft through online interaction with the environment \citep{vpt,dreamerv3}, deploying partially-trained policies is often infeasible for practical applications, such as physical robots, because ensuring safety, resetting the scene, and providing rewards in real time is difficult.
Instead, we focus on the challenge of learning purely offline from a fixed experience dataset.
We only use the VPT contractor dataset \citep{vpt}---which contains 2.5K hours of videos, actions, and event annotations---without allowing the agent to interact with the environment for learning, and compare to baselines in this offline setting.
We follow the VPT evaluation protocol of raw pixel inputs and low-level mouse and keyboard actions, requiring crafting through the in-game user interface.
Episodes last 60 minutes, starting from an empty inventory in randomly generated Minecraft worlds.

\paragraph{Implementation}
\method learns a single transformer that predicts inputs, actions, rewards, and values.
To build a steerable agent, we opt for a multi-task setting and condition the actions, rewards, and values on task embeddings.
We annotate the tasks and their sparse binary rewards using the existing events in the VPT dataset.
\Cref{tab:mc_tasks} lists the 20 tasks and \cref{tab:mc_ladder} shows the linear prompt sequence that guides the agent to reach diamonds during evaluation in the environment.
To amplify the signal in the dataset during behavior cloning, reward modeling, and reinforcement learning, we use data mixture of 50\% uniform sequences and 50\% relevant sequences that accomplish one of the tasks.
The behavioral cloning loss is applied only on the relevant fraction, while the dynamics loss is applied only on the uniform sequences to avoid optimistic generations.
We use one-hot task indicators but text embeddings could easily be used.
We represent keyboard actions as 23 binary distributions and mouse actions as a categorical with 121 classes using foveated discretization \citep{vpt}.

\pagebreak
We compare the following agents:

\begin{itemize}
\item\textbf{VPT (finetuned)}\quad
The strongest Minecraft agent for mouse and keyboard control in the literature\citep{vpt}.
The VPT paper presents two unconditioned behavioral cloning policies in the offline setting, both trained on 270K hours of synthetically annotated YouTube gameplay videos, and one further finetuned on a filtered subset of ``early game'' data.
We use the finetuned policy because it significantly outperforms the pretraining policy.

\item\textbf{BC (notask)}\quad
Behavioral cloning from scratch using multi-token prediction (MTP), without task conditioning.
While VPT uses the contractor data to train an action labeler and trains the policy on annotated YouTube videos, our BC agent trains directly and only on the relevant subset of the contractor actions.
This agent is not task conditioned, making it directly comparable to VPT.

\item\textbf{BC}\quad
Behavioral cloning from scratch on the filtered contractor dataset with task conditioning.
This agent is trained on the same filtered contractor dataset as BC (notask) but the additional task input makes it steerable.
When evaluating in the environment, the prompt sequence guides the agent through intermediate tasks towards mining diamonds.

\item\textbf{VLA (Gemma 3)}\quad
Following the VLA recipe \citep{kim2024openvla,intelligence2025pi05}, we train a behavioral cloning policy by finetuning the vision-language model Gemma 3\citep{team2025gemma3} on the relevant sequences using MTP.
Gemma 3 has been pretrained using substantially more compute and data than our other models, including native pretraining on images for visual perception, making it a strong baseline.

\item\textbf{WM+BC}\quad
The behavioral cloning policy of \method, before applying imagination training.
This policy is initialized from the world model pretrained on the full contractor dataset.
It then undergoes agent finetuning using behavioral cloning, reward modeling, and dynamics losses.

\item\textbf{Imagination RL}\quad
The full \method agent produced by finetuning the WM+BC agent through reinforcement learning in imagination, which we refer to as imagination training.
Despite performing on-policy reinforcement learning inside of the world model, there is no actual environment interaction, making it an offline method.
\end{itemize}

\begin{figure}[t!]
\newcommand{\inlineCheck}{\,\smash{\raisebox{-.3ex}{\includegraphics[height=2ex]{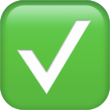}}}\,}
\newcommand{\inlineCross}{\,\smash{\raisebox{-.3ex}{\includegraphics[height=2ex]{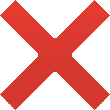}}}\,}
\vspace*{-1\baselineskip}
\centering
\includegraphics[width=\linewidth]{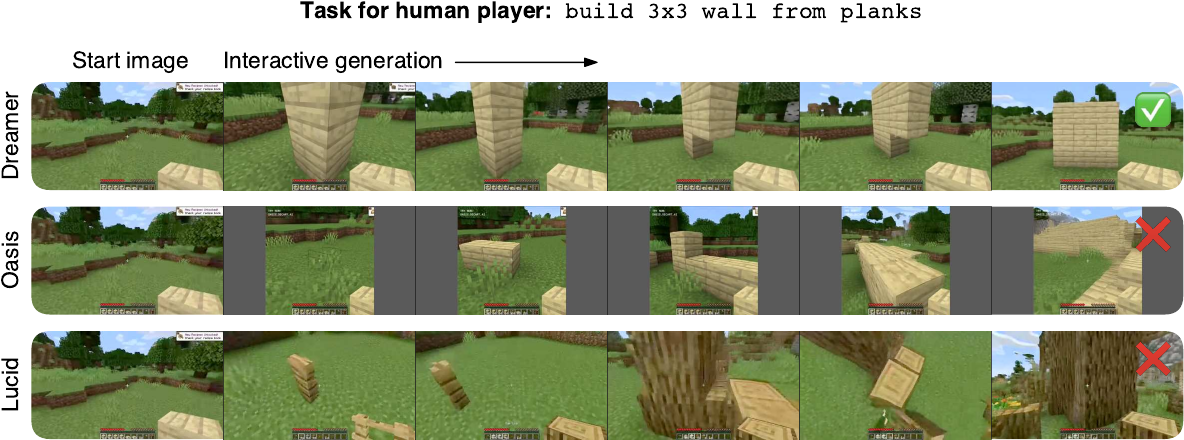}
\caption{Human interaction. A human player counterfactually interacts with the world model in real time via mouse and keyboard to perform the same task from the same initial image.
\method is the first world model to accurately predict the object interactions and game mechanics of placing the blocks in the correct shape.
In contrast, previous Minecraft world models degrade visually, change the held item, and hallucinate structures that the player never built.
The \method world model allows the player to accomplish the task (\inlineCheck) while Oasis and Lucid do not (\inlineCross).
}
\label{fig:wmtask}
\end{figure}

\paragraph{Agent performance}
\Cref{fig:rl} compares the agent performance on the diamond task.
We report the success rates for several relevant items leading up to diamonds that are listed in \cref{tab:mc_items}.
VPT (finetuned) progresses up to sticks, which it achieves 53\% of the time.
It also collects a small amount of stone, iron ore, and iron ingots through edge cases, such as exploding Creeper mobs and loot chests.
Using the contractor actions directly instead of annotating YouTube videos, our modern BC baseline achieves higher performance than VPT (finetuned).
VLA (Gemma 3) shows that initializing the policy from a pretrained models offers significant benefits, progressing up to the iron pickaxe with a success rate of 11\%.
\method achieves high success rates of over 90\% up to the stone pickaxe, a success rate of 29\% for the iron pickaxe, and obtains diamonds in 0.7\% of episodes.
Imagination training shows stronger improvements over the behavior cloning agents the more challenging the milestone is.
\Cref{fig:rlabl} compares additional agents on the diamond task, showing that the world model representations outperform the general representations of Gemma 3 for behavioral cloning.
This indicates that video prediction implicitly learns an understanding of the world that is also useful for decision making.
Finally, imagination training consistently improves not only the success rates but also makes the policy more efficient so that it reaches the milestones faster.

\subsection{Human Interaction}
\label{sec:interact}

To evaluate its ability to predict complex interactions, we train \method on the Minecraft VPT dataset \citep{vpt} and compare its generations to previous world models on this dataset.
For this evaluation, a human player tries to accomplish tasks by playing inside of the world model, as shown in \cref{fig:wmtask}.
The human player receives the task description and the world model is initialized to a start frame for the task.
We select a diverse set of tasks that cover a broad range of object interactions and game mechanics.
The tasks include digging a pit, building a wall, chopping a tree, placing and riding a boat, looking away and back at objects, interacting with crafting benches and furnaces, and more.
We compare \method to the world models Oasis \citep{oasis}, Lucid-v1 \citep{lucidv1}, and MineWorld \citep{mineworld}.
Note that we cannot directly compare to Genie 3 \citep{genie3}, because it only supports camera actions and one generic ``interact'' button, whereas Minecraft requires a more general mouse and keyboard action space.
\Cref{tab:inference} summarizes the compared models.
Full results are given in \cref{fig:wmtasks_ours,fig:wmtasks_oasis,fig:wmtasks_lucid}.

\begin{table}[t!]
\centering
\begin{mytabular}{
  colspec = {| L{8em} | C{5em} C{5em} C{5em} C{5em} C{5em} |},
  row{1} = {font=\bfseries},
  stretch=0.9,
}

\toprule
Model & Parameters & Resolution & Context & FPS & Success \\
\midrule
MineWorld       & 1.2B & 384$\times$224 & 0.8s & \o\o2 & --- \\
Lucid-v1        & 1.1B & 640$\times$360 & 1.0s &  \o44 & 0/16 \\
Oasis (small)   & 500M & 640$\times$360 & 1.6s &  \o20 & 0/16 \\
Oasis (large)   & --- & 360$\times$360 & 1.6s &  \o\o\llap{$\sim$}5 & 5/16 \\
\midrule
\method         & 2B & 640$\times$360 & 9.6s &  \o21 & \textbf{14/16}  \\
\bottomrule

\end{mytabular}
\caption{
Comparison of Minecraft world models.
\method is the first world model to accurately simulate a wide range of object interactions and game mechanics in Minecraft.
Moreover, \method pushes the limits of context length compared to previous models by 6$\times$, while maintaining real-time interactive inference.
We measure the inference speed of each model on a single H100 GPU, and translate the inference speed for the proprietary large Oasis model based on public information.
The Minecraft dataset is recorded at 20 FPS, matching the update rate of the game.
}
\label{tab:inference}
\end{table}

\pagebreak
\paragraph{Inference speed}

We measure the inference speed of each model on  a single H100 GPU.
\method and Lucid-v1 achieve real-time interactive inference by exceeding the 20 FPS of the Minecraft physics engine\footnote{\url{https://minecraft.fandom.com/wiki/Tick}} and the VPT dataset \citep{vpt}.
\method has a substantially longer context of 9.6 seconds compared to the 0.8--1.6 seconds of prior models.
Oasis comes in two sizes, a 500M parameter version with open weights and a larger model of unknown size that is playable on the project website.
The small model achieves 20 FPS on one H100.
The large model is hosted on multiple H100s for interaction online and we estimate its inference speed on a single H100 around 5 FPS based on public information.
MineWorld achieves 2 FPS with their parallel decoding approach and is even slower without.
Additionally, parallel decoding requires knowing actions in advance, which is also required in the provided user interface.
Thus, it does not support real-time interactions and we cannot evaluate it on our tasks, which require many hundreds of actions.

\begin{figure}[t!]
\centering
\vspace*{-1ex}
\includegraphics[width=\linewidth]{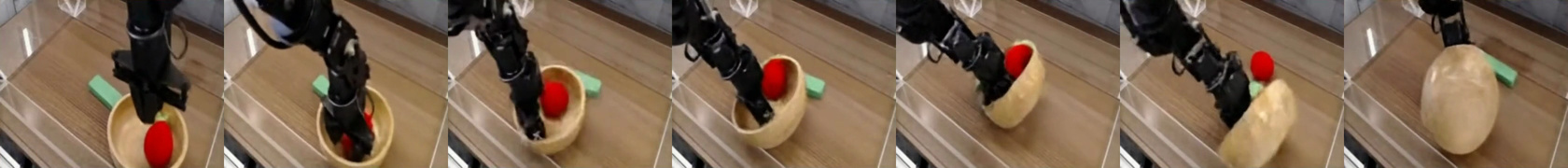} \\[1ex]
\includegraphics[width=\linewidth]{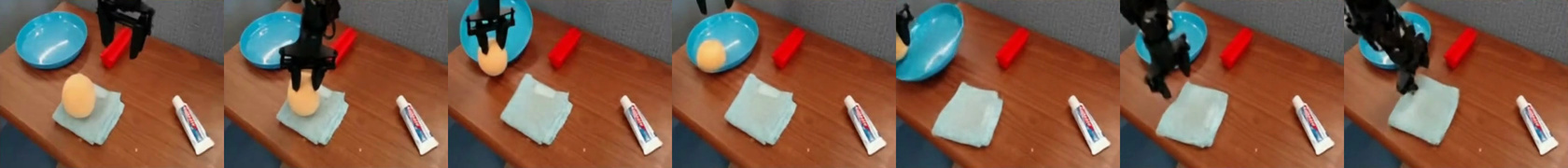} \\[1ex]
\includegraphics[width=\linewidth]{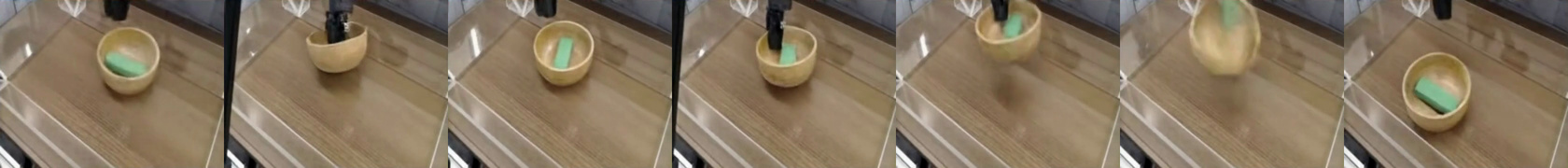}
\caption{
Robotics generations for counterfactual actions.
\method learns an accurate real-time simulator of the environment, allowing human operators to control the imagined robot to pick up objects, flip over a bowl, press a ball onto a plate, move a towel, and throw a bowl.
}
\label{fig:robotics}
\end{figure}

\enlargethispage{2\baselineskip}
\paragraph{Complex interactions}

Video generations of human players attempting all tasks inside of the world models are shown in \cref{fig:wmtasks_ours,fig:wmtasks_oasis,fig:wmtasks_lucid}.
Lucid-v1 does not allow completing the tasks, with generations diverging or object interactions being ignored.
The large Oasis model allows performing 5 out of 16, such as placing torches, filling a window with glass panes, and opening a door.
However, it fails at building tasks because after placing a few blocks, it quickly hallucinates large structures into the world.
This ``autocompletion'' failure mode reflects a lack of understanding of the game mechanics.
We did not evaluate MineWorld because of its lack of interactive inference.
\method achieves 14 out of 16 tasks, accurately generating complex interactions and game mechanics such as switching items, placing and breaking blocks, fighting monsters, placing and riding boats, entering portals, and more.
Its temporal consistency is limited to a context of 9.6 seconds, albeit substantially longer than previous models.
While it correctly generates the interfaces for inventory, crafting, and furnaces and predicts most mouse movement, inventory items are sometimes unclear or change over time, leaving room for future improvements.

As a step towards testing the applicability of \method to real world video, we also train the world model on a robotics dataset\citep{soar}.
In \cref{fig:robotics}, we observe accurate physics and counterfactual interactions with real world objects, overcoming the causal confusion of existing video models.
Additional details are included in the supplementary material.

\subsection{Action Generalization}
\label{sec:actgen}

One promise of world models is to leverage diverse unlabeled videos to teach agents about the world.
For example, a world model could learn general physics and object interactions from web videos where actions are not available.
In this section, we investigate the amount of paired videos with actions that are needed for grounding an embodiment into the \method world model.
Intuitively, the world model has to learn a broader distribution of possible outcomes when actions are missing, and can narrow the distribution down when actions are provided.
Moreover, we investigate how well the action conditioning generalizes, not just within the same distribution, but also out of distribution to parts of the world specifically held out.
To measure the accuracy of the action conditioning, we compare action-conditioned multi-step generations to ground truth videos on the holdout set.
We report PSNR and SSIM for 16-step generations given 320 frames of context.

\begin{figure}[t!]
\centering
\includegraphics[width=\linewidth]{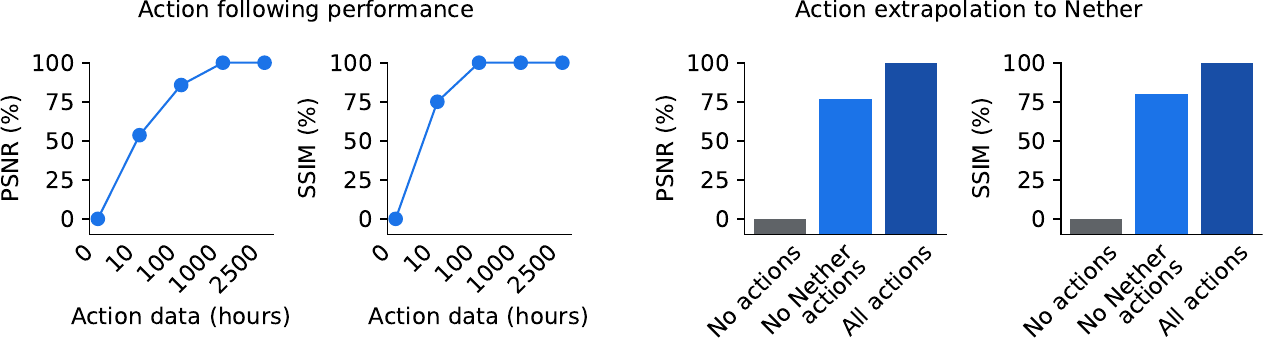}
\caption{
Action generalization.
\textbf{(left)}
\method learns accurate action conditioning from 2500 hours of video with only 100 hours of paired actions.
It achieves over 80\% of the action-conditioned generation accuracy, normalized within the range of training without any actions and using all actions.
\textbf{(right)}
When learned with only actions of the Minecraft Overworld, the action conditioning generalizes to the Nether and End dimensions of the game that are only seen in unlabeled videos.
These environments are distinct from the Overworld in their textures, blocks, and items.
The results indicate that \method learns action conditioning from small amounts of action data that generalize broadly, paving the way toward learning simulators from diverse unlabeled web videos.
}
\label{fig:actgen}
\end{figure}

\paragraph{Amount of actions}
To understand the amount of video with actions needed to learn action conditioning, we train \method on all 2541 hours of videos in the VPT dataset, but only provide actions for a small subset of the videos.
When actions are unavailable, the dynamics model is conditioned on a learned embedding, as described in \cref{sec:dynamics}.
We train \method models with 0, 10, 100, 1000, and 2541 hours of actions.
The available actions are the first in sequential order of the dataset, corresponding to fewer unique worlds and players than random shuffling would yield.
\Cref{fig:actgen} show the quality of the action conditioning compared to training with no actions at all to training with all actions.
With only 10 hours of actions, \method achieves 53\% PSNR and 75\% SSIM compared to a model trained with all actions.
With 100 hours of actions, the performance increases further to 85\% PSNR and 100\% SSIM.
This result demonstrates that world models absorb the majority of their knowledge from unlabeled videos, and require only a small amount of actions.

\paragraph{Action extrapolation}

In principle, world models may not only learn action ground from a few actions but also generalize their action conditioning to completely new scenarios.
In the future, this could allow world models to absorb general knowledge from diverse web videos to simulate agents in diverse environments.
We perform, to the best of our knowledge, the first controlled evaluation of this hypothesis.
For this, we carefully split the VPT dataset into one portion that only contains videos of the Overworld and another portion that only contains the other two game dimensions, the Nether and End.
Where the Overworld contains forests, deserts, oceans, and more, the Nether and End feature substantially different and unique visuals.
The Nether is an underworld filled with lava and red blocks and the End is filled with yellow blocks and black towers unseen in the Overworld.
We train \method on videos of both datasets but only provide actions for the Overworld.
We then perform an action-conditioned evaluation of the resulting model on the Nether and End, for which it has never seen any actions.
In a prior experiment, we observed that training only on the Overworld portion without any Nether videos results in poor generation scores for Nether start frames.
\Cref{fig:actgen} reports the relative performance compared to training without any or with all actions.
Surprisingly, the world model achieves 76\% of the PSNR and 80\% of the SSIM of the model trained with all actions.
This demonstrates that the action conditioning of world models can generalize to scenarios known only from unlabeled videos.

\subsection{Model Design}
\label{sec:modeldesign}

\begin{table}[tb!]
\centering
\vspace*{-2ex}
\begin{mytabular}{
  colspec = {| L{15em} | C{5.5em} C{5.5em} C{5.5em} |},
  row{1} = {font=\bfseries},
  stretch=0.9,
}

\toprule
Model & Train step seconds & Inference FPS (\rlap{$\uparrow$)} & Quality FVD (\rlap{$\downarrow$)} \\
\midrule

Diffusion Forcing Transformer               & 9.8 & \o0.8 &  306 \\  
+ Fewer sampling steps ($K=4$)              & 9.8 & \o9.1 &  875 \\  
+ Shortcut model                            & 9.8 & \o9.1 &  329 \\  
+ X-Prediction                              & 9.8 & \o9.1 &  326 \\  
+ X-Loss                                    & 9.8 & \o9.1 &  151 \\  
+ Ramp weight                               & 9.8 & \o9.1 &  102 \\  
+ Alternating batch lengths                 & 1.5 & \o9.1 & \o80 \\  
+ Long context every 4 layers               & 0.6 &  18.9 & \o70 \\  
+ GQA                                       & 0.5 &  23.2 & \o71 \\  
+ Time factorized long context              & 0.4 &  30.1 & \o91 \\  
+ Register tokens                           & 0.5 &  28.9 & \o91 \\  
+ More spatial tokens ($N_\mathrm{z}=128$)  & 0.8 &  25.7 & \o66 \\  
+ More spatial tokens ($N_\mathrm{z}=256$)  & 1.7 &  21.4 & \o57 \\  
\bottomrule


\end{mytabular}
\caption{
Cascade of model design choices.
\method is based on a shortcut forcing objective and an efficient transformer architecture, combining a range of known techniques to achieve accurate and fast interleaved generation.
Starting from a naive diffusion forcing transformer with $N_\mathrm{z}=64$ spatial tokens and $K=64$ sampling steps, we apply the objective and architecture modifications, and increase the number of spatial tokens once feasible.
Inference speed measured on a single H100 GPU.
The resulting world model achieves high model capacity and inference efficiency.
}
\label{tab:ablations}
\end{table}

World models require high model capacity to predict complex object interactions and fast inference to support imagination training and human interaction for inspection.
Moreover, interactive inference prompts different design choices compared to typical video models to enable fast generation of individual frames and prevent accumulating errors.
In this section, we ablate the objective and architecture decisions of \method by applying a cascade of improvements to a naive diffusion forcing transformer baseline.
To evaluate each model, we train for 48 hours and then generate 1024 videos of 384 frames (\~20 seconds) each without any context, with interactive actions chosen by a fixed behavioral cloning policy.
We then split the resulting videos into 16 frame chunks to compute the Frech\'et Video Distance (FVD) \citep{fvd} to the holdout dataset.

\pagebreak
\begin{wrapfigure}[20]{r}{0.4\textwidth}
\includegraphics[width=\linewidth]{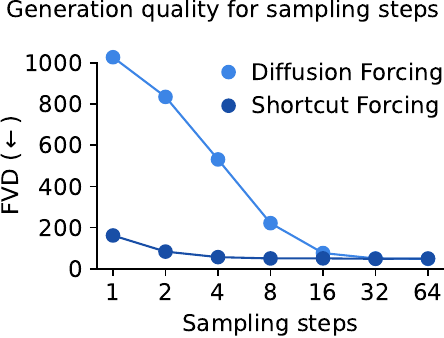}
\caption{
Generation quality of shortcut forcing compared to diffusion forcing.
Shortcut forcing with only 4 sampling steps approaches the quality of diffusion forcing with 64 steps, resulting in 16$\times$ faster generations.
}
\label{fig:nfe}
\end{wrapfigure}%
\Cref{tab:ablations} shows the progression of models and their generation quality and speed.
We start with a standard block-causal transformer with dense attention and diffusion forcing with velocity parameterization.
We target 20 FPS interactive inference on a single H100 GPU, matching the tick rate of Minecraft and the framerate of the VPT dataset.
With 64 sampling steps per frame, the baseline falls short of real-time generations with only 0.8 FPS on one H100 GPU, while 4 sampling steps achieve 9.1 FPS but result in poor quality.
Shortcut models \citep{shortcut} nearly recover the original visual quality with only 4 sampling steps.
We then parameterize the model to make x-space predictions, compute the loss in x-space, and apply the ramp loss weight.
These changes substantially improve generation quality over traditional v-space prediction, especially for long generations.
We hypothesize that the prediction targets in x-space are more structured and thus reduce the risk of high-frequency errors that accumulate over time.
\Cref{fig:nfe} compares the visual quality of shortcut forcing to diffusion forcing---both using the x-space loss with ramp weight---for a wider range of sampling steps.

Training on alternating batch lengths is similar to progressive training and speeds up learning while allowing to generate long videos for inspection throughout training.
Using temporal attention only every 4 layers not only speeds up training and inference \citep{llama4} but also improves generation quality, possibly because of the inductive bias of spatial attention that focuses computation on the current frame.
GQA further accelerates generations without degrading performance.
Switching the long context layers from dense to time-only attention accelerates inference at a mild quality cost, and prepares us to increase the overall number of tokens subsequently.
While the register tokens do not improve FVD measurably, we qualitatively notice that they improve temporal consistency.
After these changes, training and inference is fast enough to increase model capacity through more spatial tokens, improving predictions of complex interactions.
The complete model achieves an FVD of 57 compared to 306 for the naive diffusion forcing transformer baseline and 124 for the complete architecture with v-space prediction and losses.

\vfill
\begin{table}[h!]
\centering
\begin{mytabular}{
  colspec = {| L{6em} | C{5em} C{9em} | C{3.5em} C{3.5em} C{3.5em} |},
  row{1} = {font=\bfseries},
  stretch = 0.9,
}

\toprule
\SetCell[r=2]{m}{Agent} &
\SetCell[r=2]{m}{Inputs} &
\SetCell[r=2]{m}{Actions} &
\SetCell[c=3]{c}{Data (hours)} \\
& & & Offline & Web & Online \\

\midrule
{Dreamer 3} & {64$\times$64,\\inventory} & {keyboard, camera,\\ abstract crafting} & --- & --- & 1.4K \\[1ex]
VPT (RL) & 128$\times$128 & {keyboard, mouse} & 2.5K & 270K & 194K \\[1ex]
VPT (BC) & 128$\times$128 & {keyboard, mouse} & 2.5K & 270K & --- \\[1ex]
\method & 360$\times$640 & {keyboard, mouse} & 2.5K & --- & --- \\
\bottomrule

\end{mytabular}
\caption{
Comparison of experimental setups for different Minecraft agents.
\method learns purely from offline experience and requires 100$\times$ less data than previous keyboard and mouse agents.
See \cref{fig:inputs} for examples of input images.
}
\label{tab:setups}
\end{table}

\pagebreak
\section{Related Work}

\paragraph{Minecraft agents}

Obtaining diamonds in the video game Minecraft has been a focus of intelligent agent research \citep{johnson2016malmo,guss2019minerl,kanervisto2022minerlcomp,kanitscheider2021minecraftcurriculum}, a long-horizon task that requires gathering resources and crafting tools through thousands of low-level actions in complex procedurally generated 3D worlds.
Different setups have been used in the literature, with \cref{tab:setups} summarizing the ones relevant to our work.
VPT \citep{vpt} collects 2.5K hours of contractor gameplay and uses it to annotate 270K hours of web videos with synthetic mouse and keyboard actions.
Behavioral cloning on this large offline dataset and targeted finetuning on ``early-game'' data yields a policy that occasionally obtains wooden pickaxes.
Following up with 194K hours of online reinforcement learning results in a policy that obtains diamonds and diamond pickaxes.
Dreamer 3 \citep{dreamerv3} learns to collect diamonds from scratch from 1.4K hours of online interaction, without any human data.
It uses the MineRL competition action space that includes abstract crafting actions.
Other works have explored novel algorithms for easier tasks\citep{lifshitz2023steve1,cai2023minecraftgroot,zhou2024minedreamer,nieto2021minecraftskills}.
In this paper, we train an agent to achieve diamonds purely from the 2.5K hour contractor dataset, without any online interaction.
We use low-level mouse and keyboard actions and high-resolution image inputs.

\paragraph{World model agents}

Learning behaviors based on learned models of the environment has been explored for a long time \citep{sutton1991dyna,deisenroth2011pilco,watter2015e2c}.
Visual Foresight \citep{finn2017visualforesight}, the work by \citet{ha2018worldmodels}, and PlaNet \citep{hafner2018planet} achieved world models accurate enough for planning from pixels.
With Dreamer, world models have become the state-of-the-art approach for solving control problems with high-dimensional inputs, outperforming model-free reinforcement learning in robustness, efficiency, and final performance \citep{hafner2019dreamer,hafner2020dreamerv2,dreamerv3}.
World models based on transformers or diffusion objectives have demonstrated high data efficiency for discrete control \citep{micheli2022iris,robine2023twm,storm}.
However, these world models have been limited in model capacity, restricting their applicability to relatively simple simulated environments.

\paragraph{Scalable world models}

Larger world models have been shown to simulate more complex data distributions \citep{yu2025survey}.
World models like Genie 3 generate highly diverse scenes and simulate camera movement and simple interactions\citep{genie3,he2025matrix,sun2025virtual,team2025yan}.
PlayerOne\citep{tu2025playerone} and PEVA\citep{bai2025whole} condition on more detailed human movement.
Oasis \citep{oasis}, Lucid \citep{lucidv1}, and MineWorld \cite{mineworld} learn Minecraft simulators from mouse and keyboard inputs.
Whereas Oasis captures simple game mechanics and achieves real-time inference on specialized hardware, Lucid's predictions diverge quickly and MineWorld is slower than real time.
GameNGen \citep{valevski2024diffusion} finetunes Stable Diffusion into a playable simulator of a level of the game Doom.
DIAMOND \citep{alonso2024diffusion} learns a simulator of a level of the game CS GO, achieving short-term predictions.
GAIA \citep{hu2023gaia,russell2025gaia} generates driving scenarios from real world data.
However, these world models still struggle to predict complex object interactions precisely enough for imagination training.

\paragraph{Fast generation}

Enabling fast and accurate generations has been a long-standing challenge in generating modeling.
MaskGit generates discrete tokens in parallel to accelerate sampling over autoregressive models \citep{chang2022maskgit}.
Diffusion models are often distilled to enable sampling using fewer forward passes \citep{salimans2022progressive,kodaira2025streamdit}, which requires two training phases.
Consistency models learn straight paths to enable fast image generation but require careful schedules \citep{song2023consistency,song2023improved,lu2024simplifying}, with successful applications to video generation \citep{wang2023videolcm}.
Shortcut models \citep{shortcut} condition flows on both noise level and step size, producing fast generative models in one training phase and without schedules.
Mean Flow \citep{geng2025meanflow} extends the idea of conditioning on the step size to a continuous time formulation.
Efficient and sparse architectures are used for video generation because of the large number of tokens \citep{axial,zhao2024real,zhang2025vsa}.
These models do not support interactive inference and many of the techniques are complimentary to our work.

\section{Discussion}

We present \method, a scalable agent that learns to solve challenging control tasks by imagination training inside of a fast and accurate world model.
\method is the first agent to obtain diamonds in Minecraft purely from offline data, without online interaction.
This achievement demonstrates its learning successful long-horizon strategies in complex environments.
Learning purely from offline datasets enables applications where online interaction is impractical or unsafe.

The world model of \method rests on a shortcut forcing objective and an efficient transformer architecture to predict complex object interactions while supporting interactive inference in real time on a single GPU.
We demonstrate that it significantly outperforms previous world models on Minecraft, where it accurately predicts a wide range of game mechanics from mouse and keyboard inputs.
However, the world model is far from a full clone of the game, especially due to short memory and imprecise inventory predictions, leaving Minecraft as an ideal benchmark for future world model and agent research.
We also show that \method learns accurate action conditioning from only a small amount of videos with actions, allowing it to absorb the majority of its knowledge from diverse unlabeled videos.

Promising future directions include pretraining on general internet videos, integrating long-term memory into the world model and agent, incorporating language understanding, leveraging small amounts of corrective online data, and automatically discovering goals to break up long tasks.
\method offers a reliable and performant starting point for these explorations.

\clearpage
\begin{hyphenrules}{nohyphenation}
\setlength{\bibsep}{.5ex plus .5ex}
\bibliography{references}

\begin{thebibliography}{84}
\providecommand{\natexlab}[1]{#1}
\providecommand{\url}[1]{\texttt{#1}}
\expandafter\ifx\csname urlstyle\endcsname\relax
  \providecommand{\doi}[1]{doi: #1}\else
  \providecommand{\doi}{doi: \begingroup \urlstyle{rm}\Url}\fi

\bibitem[Hafner et~al.(2025)Hafner, Pasukonis, Ba, and Lillicrap]{dreamerv3}
Danijar Hafner, Jurgis Pasukonis, Jimmy Ba, and Timothy Lillicrap.
\newblock Mastering diverse control tasks through world models.
\newblock \emph{Nature}, pages 1--7, 2025.

\bibitem[Wu et~al.(2023)Wu, Escontrela, Hafner, Abbeel, and
  Goldberg]{wu2023daydreamer}
Philipp Wu, Alejandro Escontrela, Danijar Hafner, Pieter Abbeel, and Ken
  Goldberg.
\newblock Daydreamer: World models for physical robot learning.
\newblock In \emph{Conference on robot learning}, pages 2226--2240. PMLR, 2023.

\bibitem[Hansen et~al.(2023)Hansen, Su, and Wang]{hansen2023tdtmpc2}
Nicklas Hansen, Hao Su, and Xiaolong Wang.
\newblock Td-mpc2: Scalable, robust world models for continuous control.
\newblock \emph{arXiv preprint arXiv:2310.16828}, 2023.

\bibitem[Alonso et~al.(2024)Alonso, Jelley, Micheli, Kanervisto, Storkey,
  Pearce, and Fleuret]{alonso2024diffusion}
Eloi Alonso, Adam Jelley, Vincent Micheli, Anssi Kanervisto, Amos~J Storkey,
  Tim Pearce, and Fran{\c{c}}ois Fleuret.
\newblock Diffusion for world modeling: Visual details matter in atari.
\newblock \emph{Advances in Neural Information Processing Systems},
  37:\penalty0 58757--58791, 2024.

\bibitem[Schrittwieser et~al.(2019)Schrittwieser, Antonoglou, Hubert, Simonyan,
  Sifre, Schmitt, Guez, Lockhart, Hassabis, Graepel,
  et~al.]{schrittwieser2019muzero}
Julian Schrittwieser, Ioannis Antonoglou, Thomas Hubert, Karen Simonyan,
  Laurent Sifre, Simon Schmitt, Arthur Guez, Edward Lockhart, Demis Hassabis,
  Thore Graepel, et~al.
\newblock Mastering atari, go, chess and shogi by planning with a learned
  model.
\newblock \emph{arXiv preprint arXiv:1911.08265}, 2019.

\bibitem[Hessel et~al.(2021)Hessel, Danihelka, Viola, Guez, Schmitt, Sifre,
  Weber, Silver, and Van~Hasselt]{hessel2021muesli}
Matteo Hessel, Ivo Danihelka, Fabio Viola, Arthur Guez, Simon Schmitt, Laurent
  Sifre, Theophane Weber, David Silver, and Hado Van~Hasselt.
\newblock Muesli: Combining improvements in policy optimization.
\newblock In \emph{International Conference on Machine Learning}, pages
  4214--4226. PMLR, 2021.

\bibitem[Ball et~al.(2025)Ball, Bauer, Belletti, Brownfield, Ephrat, Fruchter,
  Gupta, Holsheimer, Holynski, Hron, Kaplanis, Limont, McGill, Oliveira,
  Parker-Holder, Perbet, Scully, Shar, Spencer, Tov, Villegas, Wang, Yung,
  Baetu, Berbel, Bridson, Bruce, Buttimore, Chakera, Chandra, Collins, Cullum,
  Damoc, Dasagi, Gazeau, Gbadamosi, Han, Hirst, Kachra, Kerley, Kjems,
  Knoepfel, Koriakin, Lo, Lu, Mehring, Moufarek, Nandwani, Oliveira, Pardo,
  Park, Pierson, Poole, Ran, Salimans, Sanchez, Saprykin, Shen, Sidhwani,
  Smith, Stanton, Tomlinson, Vijaykumar, Wang, Wingfield, Wong, Xu, Yew, Young,
  Zubov, Eck, Erhan, Kavukcuoglu, Hassabis, Gharamani, Hadsell, van~den Oord,
  Mosseri, Bolton, Singh, and Rockt{\"a}schel]{genie3}
Philip~J. Ball, Jakob Bauer, Frank Belletti, Bethanie Brownfield, Ariel Ephrat,
  Shlomi Fruchter, Agrim Gupta, Kristian Holsheimer, Aleksander Holynski, Jiri
  Hron, Christos Kaplanis, Marjorie Limont, Matt McGill, Yanko Oliveira, Jack
  Parker-Holder, Frank Perbet, Guy Scully, Jeremy Shar, Stephen Spencer, Omer
  Tov, Ruben Villegas, Emma Wang, Jessica Yung, Cip Baetu, Jordi Berbel, David
  Bridson, Jake Bruce, Gavin Buttimore, Sarah Chakera, Bilva Chandra, Paul
  Collins, Alex Cullum, Bogdan Damoc, Vibha Dasagi, Maxime Gazeau, Charles
  Gbadamosi, Woohyun Han, Ed~Hirst, Ashyana Kachra, Lucie Kerley, Kristian
  Kjems, Eva Knoepfel, Vika Koriakin, Jessica Lo, Cong Lu, Zeb Mehring, Alex
  Moufarek, Henna Nandwani, Valeria Oliveira, Fabio Pardo, Jane Park, Andrew
  Pierson, Ben Poole, Helen Ran, Tim Salimans, Manuel Sanchez, Igor Saprykin,
  Amy Shen, Sailesh Sidhwani, Duncan Smith, Joe Stanton, Hamish Tomlinson,
  Dimple Vijaykumar, Luyu Wang, Piers Wingfield, Nat Wong, Keyang Xu,
  Christopher Yew, Nick Young, Vadim Zubov, Douglas Eck, Dumitru Erhan, Koray
  Kavukcuoglu, Demis Hassabis, Zoubin Gharamani, Raia Hadsell, A{\"a}ron
  van~den Oord, Inbar Mosseri, Adrian Bolton, Satinder Singh, and Tim
  Rockt{\"a}schel.
\newblock Genie 3: A new frontier for world models.
\newblock
  \url{https://deepmind.google/discover/blog/genie-3-a-new-frontier-for-world-models/},
  2025.

\bibitem[Tu et~al.(2025)Tu, Luo, Chen, Bai, Wang, and Zhao]{tu2025playerone}
Yuanpeng Tu, Hao Luo, Xi~Chen, Xiang Bai, Fan Wang, and Hengshuang Zhao.
\newblock Playerone: Egocentric world simulator.
\newblock \emph{arXiv preprint arXiv:2506.09995}, 2025.

\bibitem[He et~al.(2025)He, Peng, Liu, Wang, Zhang, Cui, Kang, Jiang, An, Ren,
  et~al.]{he2025matrix}
Xianglong He, Chunli Peng, Zexiang Liu, Boyang Wang, Yifan Zhang, Qi~Cui, Fei
  Kang, Biao Jiang, Mengyin An, Yangyang Ren, et~al.
\newblock Matrix-game 2.0: An open-source, real-time, and streaming interactive
  world model.
\newblock \emph{arXiv preprint arXiv:2508.13009}, 2025.

\bibitem[Sun et~al.(2025)Sun, Wei, Zhao, Chen, Chen, Zhang, Zhang, and
  Lu]{sun2025virtual}
Wenqiang Sun, Fangyun Wei, Jinjing Zhao, Xi~Chen, Zilong Chen, Hongyang Zhang,
  Jun Zhang, and Yan Lu.
\newblock From virtual games to real-world play.
\newblock \emph{arXiv preprint arXiv:2506.18901}, 2025.

\bibitem[Bai et~al.(2025)Bai, Tran, Bar, LeCun, Darrell, and
  Malik]{bai2025whole}
Yutong Bai, Danny Tran, Amir Bar, Yann LeCun, Trevor Darrell, and Jitendra
  Malik.
\newblock Whole-body conditioned egocentric video prediction.
\newblock \emph{arXiv preprint arXiv:2506.21552}, 2025.

\bibitem[Team(2025)]{team2025yan}
Yan Team.
\newblock Yan: Foundational interactive video generation.
\newblock \emph{arXiv preprint arXiv:2508.08601}, 2025.

\bibitem[Peebles and Xie(2023)]{peebles2023dit}
William Peebles and Saining Xie.
\newblock Scalable diffusion models with transformers.
\newblock In \emph{Proceedings of the IEEE/CVF international conference on
  computer vision}, pages 4195--4205, 2023.

\bibitem[Chen et~al.(2024)Chen, Mart{\'\i}~Mons{\'o}, Du, Simchowitz, Tedrake,
  and Sitzmann]{diffusionforcing}
Boyuan Chen, Diego Mart{\'\i}~Mons{\'o}, Yilun Du, Max Simchowitz, Russ
  Tedrake, and Vincent Sitzmann.
\newblock Diffusion forcing: Next-token prediction meets full-sequence
  diffusion.
\newblock \emph{Advances in Neural Information Processing Systems},
  37:\penalty0 24081--24125, 2024.

\bibitem[Baker et~al.(2022)Baker, Akkaya, Zhokov, Huizinga, Tang, Ecoffet,
  Houghton, Sampedro, and Clune]{vpt}
Bowen Baker, Ilge Akkaya, Peter Zhokov, Joost Huizinga, Jie Tang, Adrien
  Ecoffet, Brandon Houghton, Raul Sampedro, and Jeff Clune.
\newblock Video pretraining (vpt): Learning to act by watching unlabeled online
  videos.
\newblock \emph{Advances in Neural Information Processing Systems},
  35:\penalty0 24639--24654, 2022.

\bibitem[Sohl-Dickstein et~al.(2015)Sohl-Dickstein, Weiss, Maheswaranathan, and
  Ganguli]{sohl2015deep}
Jascha Sohl-Dickstein, Eric Weiss, Niru Maheswaranathan, and Surya Ganguli.
\newblock Deep unsupervised learning using nonequilibrium thermodynamics.
\newblock In \emph{International conference on machine learning}, pages
  2256--2265. pmlr, 2015.

\bibitem[Ho et~al.(2020)Ho, Jain, and Abbeel]{ddpm}
Jonathan Ho, Ajay Jain, and Pieter Abbeel.
\newblock Denoising diffusion probabilistic models.
\newblock \emph{Advances in neural information processing systems},
  33:\penalty0 6840--6851, 2020.

\bibitem[Lipman et~al.(2022)Lipman, Chen, Ben-Hamu, Nickel, and
  Le]{flowmatching}
Yaron Lipman, Ricky~TQ Chen, Heli Ben-Hamu, Maximilian Nickel, and Matt Le.
\newblock Flow matching for generative modeling.
\newblock \emph{arXiv preprint arXiv:2210.02747}, 2022.

\bibitem[Liu et~al.(2022)Liu, Gong, and Liu]{rectifiedflow}
Xingchao Liu, Chengyue Gong, and Qiang Liu.
\newblock Flow straight and fast: Learning to generate and transfer data with
  rectified flow.
\newblock \emph{arXiv preprint arXiv:2209.03003}, 2022.

\bibitem[Esser et~al.(2024)Esser, Kulal, Blattmann, Entezari, M{\"u}ller,
  Saini, Levi, Lorenz, Sauer, Boesel, et~al.]{sd3}
Patrick Esser, Sumith Kulal, Andreas Blattmann, Rahim Entezari, Jonas
  M{\"u}ller, Harry Saini, Yam Levi, Dominik Lorenz, Axel Sauer, Frederic
  Boesel, et~al.
\newblock Scaling rectified flow transformers for high-resolution image
  synthesis.
\newblock In \emph{Forty-first international conference on machine learning},
  2024.

\bibitem[Frans et~al.(2024)Frans, Hafner, Levine, and Abbeel]{shortcut}
Kevin Frans, Danijar Hafner, Sergey Levine, and Pieter Abbeel.
\newblock One step diffusion via shortcut models.
\newblock \emph{arXiv preprint arXiv:2410.12557}, 2024.

\bibitem[Zhang et~al.(2018)Zhang, Isola, Efros, Shechtman, and Wang]{lpips}
Richard Zhang, Phillip Isola, Alexei~A Efros, Eli Shechtman, and Oliver Wang.
\newblock The unreasonable effectiveness of deep features as a perceptual
  metric.
\newblock In \emph{Proceedings of the IEEE conference on computer vision and
  pattern recognition}, pages 586--595, 2018.

\bibitem[He et~al.(2022)He, Chen, Xie, Li, Doll{\'a}r, and Girshick]{mae}
Kaiming He, Xinlei Chen, Saining Xie, Yanghao Li, Piotr Doll{\'a}r, and Ross
  Girshick.
\newblock Masked autoencoders are scalable vision learners.
\newblock In \emph{Proceedings of the IEEE/CVF conference on computer vision
  and pattern recognition}, pages 16000--16009, 2022.

\bibitem[Chen et~al.(2025)Chen, Han, Chen, Li, Wang, Wang, Wang, Liu, Zou, and
  Raj]{chen2025maetok}
Hao Chen, Yujin Han, Fangyi Chen, Xiang Li, Yidong Wang, Jindong Wang, Ze~Wang,
  Zicheng Liu, Difan Zou, and Bhiksha Raj.
\newblock Masked autoencoders are effective tokenizers for diffusion models.
\newblock In \emph{Forty-second International Conference on Machine Learning},
  2025.

\bibitem[Darcet et~al.(2023)Darcet, Oquab, Mairal, and Bojanowski]{vitregister}
Timoth{\'e}e Darcet, Maxime Oquab, Julien Mairal, and Piotr Bojanowski.
\newblock Vision transformers need registers.
\newblock \emph{arXiv preprint arXiv:2309.16588}, 2023.

\bibitem[Kingma and Gao(2023)]{kingma2023understanding}
Diederik Kingma and Ruiqi Gao.
\newblock Understanding diffusion objectives as the elbo with simple data
  augmentation.
\newblock \emph{Advances in Neural Information Processing Systems},
  36:\penalty0 65484--65516, 2023.

\bibitem[Gloeckle et~al.(2024)Gloeckle, Idrissi, Rozi{\`e}re, Lopez-Paz, and
  Synnaeve]{gloeckle2024mtp}
Fabian Gloeckle, Badr~Youbi Idrissi, Baptiste Rozi{\`e}re, David Lopez-Paz, and
  Gabriel Synnaeve.
\newblock Better \& faster large language models via multi-token prediction.
\newblock \emph{arXiv preprint arXiv:2404.19737}, 2024.

\bibitem[Sutton(1988)]{sutton1988td}
Richard~S Sutton.
\newblock Learning to predict by the methods of temporal differences.
\newblock \emph{Machine learning}, 3\penalty0 (1):\penalty0 9--44, 1988.

\bibitem[Abdolmaleki et~al.(2024)Abdolmaleki, Piot, Shahriari, Springenberg,
  Hertweck, Joshi, Oh, Bloesch, Lampe, Heess, et~al.]{abdolmaleki2024pmpo}
Abbas Abdolmaleki, Bilal Piot, Bobak Shahriari, Jost~Tobias Springenberg, Tim
  Hertweck, Rishabh Joshi, Junhyuk Oh, Michael Bloesch, Thomas Lampe, Nicolas
  Heess, et~al.
\newblock Preference optimization as probabilistic inference.
\newblock \emph{arXiv e-prints}, pages arXiv--2410, 2024.

\bibitem[Shannon(1948)]{shannon1948infotheory}
Claude~E Shannon.
\newblock A mathematical theory of communication.
\newblock \emph{Bell system technical journal}, 27\penalty0 (3):\penalty0
  379--423, 1948.

\bibitem[Vaswani(2017)]{vaswani2017transformer}
A~Vaswani.
\newblock Attention is all you need.
\newblock \emph{Advances in Neural Information Processing Systems}, 2017.

\bibitem[Zhang and Sennrich(2019)]{rmsnorm}
Biao Zhang and Rico Sennrich.
\newblock Root mean square layer normalization.
\newblock \emph{Advances in Neural Information Processing Systems}, 32, 2019.

\bibitem[Su et~al.(2021)Su, Zhang, Li, Zhang, and Li]{rope}
J~Su, H~Zhang, X~Li, J~Zhang, and Y~RoFormer Li.
\newblock Enhanced transformer with rotary position embedding.
\newblock In \emph{Proceedings of the 59th Annual Meeting of the Association
  for Computational Linguistics and the 11th International Joint Conference on
  Natural Language Processing (ACL-IJCNLP), Association for Computational
  Linguistics, Online}, pages 1--6, 2021.

\bibitem[Shazeer(2020)]{swiglu}
Noam Shazeer.
\newblock Glu variants improve transformer.
\newblock \emph{arXiv preprint arXiv:2002.05202}, 2020.

\bibitem[Dehghani et~al.(2023)Dehghani, Djolonga, Mustafa, Padlewski, Heek,
  Gilmer, Steiner, Caron, Geirhos, Alabdulmohsin, et~al.]{dehghani2023scaling}
Mostafa Dehghani, Josip Djolonga, Basil Mustafa, Piotr Padlewski, Jonathan
  Heek, Justin Gilmer, Andreas~Peter Steiner, Mathilde Caron, Robert Geirhos,
  Ibrahim Alabdulmohsin, et~al.
\newblock Scaling vision transformers to 22 billion parameters.
\newblock In \emph{International Conference on Machine Learning}, pages
  7480--7512. PMLR, 2023.

\bibitem[Bello et~al.(2016)Bello, Pham, Le, Norouzi, and
  Bengio]{bello2016neural}
Irwan Bello, Hieu Pham, Quoc~V Le, Mohammad Norouzi, and Samy Bengio.
\newblock Neural combinatorial optimization with reinforcement learning.
\newblock \emph{arXiv preprint arXiv:1611.09940}, 2016.

\bibitem[Team et~al.(2024)Team, Riviere, Pathak, Sessa, Hardin, Bhupatiraju,
  Hussenot, Mesnard, Shahriari, Ram{\'e}, et~al.]{gemma2}
Gemma Team, Morgane Riviere, Shreya Pathak, Pier~Giuseppe Sessa, Cassidy
  Hardin, Surya Bhupatiraju, L{\'e}onard Hussenot, Thomas Mesnard, Bobak
  Shahriari, Alexandre Ram{\'e}, et~al.
\newblock Gemma 2: Improving open language models at a practical size.
\newblock \emph{arXiv preprint arXiv:2408.00118}, 2024.

\bibitem[Ho et~al.(2019)Ho, Kalchbrenner, Weissenborn, and Salimans]{axial}
Jonathan Ho, Nal Kalchbrenner, Dirk Weissenborn, and Tim Salimans.
\newblock Axial attention in multidimensional transformers.
\newblock \emph{arXiv preprint arXiv:1912.12180}, 2019.

\bibitem[Singh(2025)]{llama4}
Ajit Singh.
\newblock Meta llama 4: The future of multimodal ai.
\newblock \emph{Available at SSRN 5208228}, 2025.

\bibitem[Ainslie et~al.(2023)Ainslie, Lee-Thorp, De~Jong, Zemlyanskiy,
  Lebr{\'o}n, and Sanghai]{gqa}
Joshua Ainslie, James Lee-Thorp, Michiel De~Jong, Yury Zemlyanskiy, Federico
  Lebr{\'o}n, and Sumit Sanghai.
\newblock Gqa: Training generalized multi-query transformer models from
  multi-head checkpoints.
\newblock \emph{arXiv preprint arXiv:2305.13245}, 2023.

\bibitem[Kim et~al.(2024)Kim, Pertsch, Karamcheti, Xiao, Balakrishna, Nair,
  Rafailov, Foster, Lam, Sanketi, et~al.]{kim2024openvla}
Moo~Jin Kim, Karl Pertsch, Siddharth Karamcheti, Ted Xiao, Ashwin Balakrishna,
  Suraj Nair, Rafael Rafailov, Ethan Foster, Grace Lam, Pannag Sanketi, et~al.
\newblock Openvla: An open-source vision-language-action model.
\newblock \emph{arXiv preprint arXiv:2406.09246}, 2024.

\bibitem[Intelligence et~al.(2025)Intelligence, Black, Brown, Darpinian,
  Dhabalia, Driess, Esmail, Equi, Finn, Fusai, et~al.]{intelligence2025pi05}
Physical Intelligence, Kevin Black, Noah Brown, James Darpinian, Karan
  Dhabalia, Danny Driess, Adnan Esmail, Michael Equi, Chelsea Finn, Niccolo
  Fusai, et~al.
\newblock pi0.5: a vision-language-action model with open-world generalization.
\newblock \emph{arXiv preprint arXiv:2504.16054}, 2025.

\bibitem[Team et~al.(2025)Team, Kamath, Ferret, Pathak, Vieillard, Merhej,
  Perrin, Matejovicova, Ram{\'e}, Rivi{\`e}re, et~al.]{team2025gemma3}
Gemma Team, Aishwarya Kamath, Johan Ferret, Shreya Pathak, Nino Vieillard,
  Ramona Merhej, Sarah Perrin, Tatiana Matejovicova, Alexandre Ram{\'e},
  Morgane Rivi{\`e}re, et~al.
\newblock Gemma 3 technical report.
\newblock \emph{arXiv preprint arXiv:2503.19786}, 2025.

\bibitem[Zhao et~al.(2023)Zhao, Gu, Varma, Luo, Huang, Xu, Wright, Shojanazeri,
  Ott, Shleifer, et~al.]{fsdp}
Yanli Zhao, Andrew Gu, Rohan Varma, Liang Luo, Chien-Chin Huang, Min Xu, Less
  Wright, Hamid Shojanazeri, Myle Ott, Sam Shleifer, et~al.
\newblock Pytorch fsdp: experiences on scaling fully sharded data parallel.
\newblock \emph{arXiv preprint arXiv:2304.11277}, 2023.

\bibitem[Rasley et~al.(2020)Rasley, Rajbhandari, Ruwase, and He]{deepspeed}
Jeff Rasley, Samyam Rajbhandari, Olatunji Ruwase, and Yuxiong He.
\newblock Deepspeed: System optimizations enable training deep learning models
  with over 100 billion parameters.
\newblock In \emph{Proceedings of the 26th ACM SIGKDD international conference
  on knowledge discovery \& data mining}, pages 3505--3506, 2020.

\bibitem[Decart and Etched(2024)]{oasis}
Decart and Etched.
\newblock Oasis: A universe in a transformer.
\newblock https://www.decart.ai/articles/oasis-interactive-ai-video-game-model,
  2024.

\bibitem[Seid and Hojel(2024)]{lucidv1}
Rami Seid and Alberto Hojel.
\newblock Lucid v1: Real-tiem latent world models.
\newblock \emph{International Journal of Current Research in Science,
  Engineering \& Technology}, 2024.

\bibitem[Guo et~al.(2025)Guo, Ye, He, Wu, Jiang, Pearce, and Bian]{mineworld}
Junliang Guo, Yang Ye, Tianyu He, Haoyu Wu, Yushu Jiang, Tim Pearce, and Jiang
  Bian.
\newblock Mineworld: a real-time and open-source interactive world model on
  minecraft.
\newblock \emph{arXiv preprint arXiv:2504.08388}, 2025.

\bibitem[Zhou et~al.(2024{\natexlab{a}})Zhou, Atreya, Lee, Walke, Mees, and
  Levine]{soar}
Zhiyuan Zhou, Pranav Atreya, Abraham Lee, Homer Walke, Oier Mees, and Sergey
  Levine.
\newblock Autonomous improvement of instruction following skills via foundation
  models.
\newblock \emph{arXiv preprint arXiv:2407.20635}, 2024{\natexlab{a}}.

\bibitem[Unterthiner et~al.(2019)Unterthiner, Van~Steenkiste, Kurach, Marinier,
  Michalski, and Gelly]{fvd}
Thomas Unterthiner, Sjoerd Van~Steenkiste, Karol Kurach, Rapha{\"e}l Marinier,
  Marcin Michalski, and Sylvain Gelly.
\newblock Fvd: A new metric for video generation.
\newblock \emph{ICLR Workshop on Deep Generative Models for Highly Structured
  Data}, 2019.

\bibitem[Johnson et~al.(2016)Johnson, Hofmann, Hutton, and
  Bignell]{johnson2016malmo}
Matthew Johnson, Katja Hofmann, Tim Hutton, and David Bignell.
\newblock The malmo platform for artificial intelligence experimentation.
\newblock In \emph{IJCAI}, pages 4246--4247. Citeseer, 2016.

\bibitem[Guss et~al.(2019)Guss, Codel, Hofmann, Houghton, Kuno, Milani,
  Mohanty, Perez~Liebana, Salakhutdinov, Topin, et~al.]{guss2019minerl}
William~H Guss, Cayden Codel, Katja Hofmann, Brandon Houghton, Noboru Kuno,
  Stephanie Milani, Sharada Mohanty, Diego Perez~Liebana, Ruslan Salakhutdinov,
  Nicholay Topin, et~al.
\newblock The minerl competition on sample efficient reinforcement learning
  using human priors.
\newblock \emph{arXiv e-prints}, pages arXiv--1904, 2019.

\bibitem[Kanervisto et~al.(2022)Kanervisto, Milani, Ramanauskas, Topin, Lin,
  Li, Shi, Ye, Fu, Yang, et~al.]{kanervisto2022minerlcomp}
Anssi Kanervisto, Stephanie Milani, Karolis Ramanauskas, Nicholay Topin,
  Zichuan Lin, Junyou Li, Jianing Shi, Deheng Ye, Qiang Fu, Wei Yang, et~al.
\newblock Minerl diamond 2021 competition: Overview, results, and lessons
  learned.
\newblock \emph{NeurIPS 2021 Competitions and Demonstrations Track}, pages
  13--28, 2022.

\bibitem[Kanitscheider et~al.(2021)Kanitscheider, Huizinga, Farhi, Guss,
  Houghton, Sampedro, Zhokhov, Baker, Ecoffet, Tang,
  et~al.]{kanitscheider2021minecraftcurriculum}
Ingmar Kanitscheider, Joost Huizinga, David Farhi, William~Hebgen Guss, Brandon
  Houghton, Raul Sampedro, Peter Zhokhov, Bowen Baker, Adrien Ecoffet, Jie
  Tang, et~al.
\newblock Multi-task curriculum learning in a complex, visual, hard-exploration
  domain: Minecraft.
\newblock \emph{arXiv preprint arXiv:2106.14876}, 2021.

\bibitem[Lifshitz et~al.(2023)Lifshitz, Paster, Chan, Ba, and
  McIlraith]{lifshitz2023steve1}
Shalev Lifshitz, Keiran Paster, Harris Chan, Jimmy Ba, and Sheila McIlraith.
\newblock Steve-1: A generative model for text-to-behavior in minecraft.
\newblock \emph{Advances in Neural Information Processing Systems},
  36:\penalty0 69900--69929, 2023.

\bibitem[Cai et~al.(2023)Cai, Zhang, Wang, Ma, Liu, and
  Liang]{cai2023minecraftgroot}
Shaofei Cai, Bowei Zhang, Zihao Wang, Xiaojian Ma, Anji Liu, and Yitao Liang.
\newblock Groot: Learning to follow instructions by watching gameplay videos.
\newblock \emph{arXiv preprint arXiv:2310.08235}, 2023.

\bibitem[Zhou et~al.(2024{\natexlab{b}})Zhou, Qin, Yin, Huang, Zhang, Sheng,
  Qiao, and Shao]{zhou2024minedreamer}
Enshen Zhou, Yiran Qin, Zhenfei Yin, Yuzhou Huang, Ruimao Zhang, Lu~Sheng,
  Yu~Qiao, and Jing Shao.
\newblock Minedreamer: Learning to follow instructions via chain-of-imagination
  for simulated-world control.
\newblock \emph{arXiv preprint arXiv:2403.12037}, 2024{\natexlab{b}}.

\bibitem[Nieto et~al.(2021)Nieto, Creus, and Giro-i
  Nieto]{nieto2021minecraftskills}
Juan~Jos{\'e} Nieto, Roger Creus, and Xavier Giro-i Nieto.
\newblock Unsupervised skill-discovery and skill-learning in minecraft.
\newblock \emph{arXiv preprint arXiv:2107.08398}, 2021.

\bibitem[Sutton(1991)]{sutton1991dyna}
Richard~S Sutton.
\newblock Dyna, an integrated architecture for learning, planning, and
  reacting.
\newblock \emph{ACM SIGART Bulletin}, 2\penalty0 (4):\penalty0 160--163, 1991.

\bibitem[Deisenroth and Rasmussen(2011)]{deisenroth2011pilco}
Marc Deisenroth and Carl~E Rasmussen.
\newblock Pilco: A model-based and data-efficient approach to policy search.
\newblock In \emph{Proceedings of the 28th International Conference on machine
  learning (ICML-11)}, pages 465--472, 2011.

\bibitem[Watter et~al.(2015)Watter, Springenberg, Boedecker, and
  Riedmiller]{watter2015e2c}
Manuel Watter, Jost Springenberg, Joschka Boedecker, and Martin Riedmiller.
\newblock Embed to control: A locally linear latent dynamics model for control
  from raw images.
\newblock In \emph{Advances in neural information processing systems}, pages
  2746--2754, 2015.

\bibitem[Finn and Levine(2017)]{finn2017visualforesight}
Chelsea Finn and Sergey Levine.
\newblock Deep visual foresight for planning robot motion.
\newblock In \emph{2017 IEEE International Conference on Robotics and
  Automation (ICRA)}, pages 2786--2793. IEEE, 2017.

\bibitem[Ha and Schmidhuber(2018)]{ha2018worldmodels}
David Ha and J{\"u}rgen Schmidhuber.
\newblock World models.
\newblock \emph{arXiv preprint arXiv:1803.10122}, 2018.

\bibitem[Hafner et~al.(2018)Hafner, Lillicrap, Fischer, Villegas, Ha, Lee, and
  Davidson]{hafner2018planet}
Danijar Hafner, Timothy Lillicrap, Ian Fischer, Ruben Villegas, David Ha,
  Honglak Lee, and James Davidson.
\newblock Learning latent dynamics for planning from pixels.
\newblock \emph{arXiv preprint arXiv:1811.04551}, 2018.

\bibitem[Hafner et~al.(2019)Hafner, Lillicrap, Ba, and
  Norouzi]{hafner2019dreamer}
Danijar Hafner, Timothy Lillicrap, Jimmy Ba, and Mohammad Norouzi.
\newblock Dream to control: Learning behaviors by latent imagination.
\newblock \emph{arXiv preprint arXiv:1912.01603}, 2019.

\bibitem[Hafner et~al.(2020)Hafner, Lillicrap, Norouzi, and
  Ba]{hafner2020dreamerv2}
Danijar Hafner, Timothy Lillicrap, Mohammad Norouzi, and Jimmy Ba.
\newblock Mastering atari with discrete world models.
\newblock \emph{arXiv preprint arXiv:2010.02193}, 2020.

\bibitem[Micheli et~al.(2022)Micheli, Alonso, and Fleuret]{micheli2022iris}
Vincent Micheli, Eloi Alonso, and Fran{\c{c}}ois Fleuret.
\newblock Transformers are sample efficient world models.
\newblock \emph{arXiv preprint arXiv:2209.00588}, 2022.

\bibitem[Robine et~al.(2023)Robine, H{\"o}ftmann, Uelwer, and
  Harmeling]{robine2023twm}
Jan Robine, Marc H{\"o}ftmann, Tobias Uelwer, and Stefan Harmeling.
\newblock Transformer-based world models are happy with 100k interactions.
\newblock \emph{arXiv preprint arXiv:2303.07109}, 2023.

\bibitem[Zhang et~al.(2023)Zhang, Wang, Sun, Yuan, and Huang]{storm}
Weipu Zhang, Gang Wang, Jian Sun, Yetian Yuan, and Gao Huang.
\newblock Storm: Efficient stochastic transformer based world models for
  reinforcement learning.
\newblock \emph{Advances in Neural Information Processing Systems},
  36:\penalty0 27147--27166, 2023.

\bibitem[Yu et~al.(2025)Yu, Qin, Che, Liu, Wang, Wan, Zhang, Gai, Chen, and
  Liu]{yu2025survey}
Jiwen Yu, Yiran Qin, Haoxuan Che, Quande Liu, Xintao Wang, Pengfei Wan,
  Di~Zhang, Kun Gai, Hao Chen, and Xihui Liu.
\newblock A survey of interactive generative video.
\newblock \emph{arXiv preprint arXiv:2504.21853}, 2025.

\bibitem[Valevski et~al.(2024)Valevski, Leviathan, Arar, and
  Fruchter]{valevski2024diffusion}
Dani Valevski, Yaniv Leviathan, Moab Arar, and Shlomi Fruchter.
\newblock Diffusion models are real-time game engines.
\newblock \emph{arXiv preprint arXiv:2408.14837}, 2024.

\bibitem[Hu et~al.(2023)Hu, Russell, Yeo, Murez, Fedoseev, Kendall, Shotton,
  and Corrado]{hu2023gaia}
Anthony Hu, Lloyd Russell, Hudson Yeo, Zak Murez, George Fedoseev, Alex
  Kendall, Jamie Shotton, and Gianluca Corrado.
\newblock Gaia-1: A generative world model for autonomous driving.
\newblock \emph{arXiv preprint arXiv:2309.17080}, 2023.

\bibitem[Russell et~al.(2025)Russell, Hu, Bertoni, Fedoseev, Shotton, Arani,
  and Corrado]{russell2025gaia}
Lloyd Russell, Anthony Hu, Lorenzo Bertoni, George Fedoseev, Jamie Shotton,
  Elahe Arani, and Gianluca Corrado.
\newblock Gaia-2: A controllable multi-view generative world model for
  autonomous driving.
\newblock \emph{arXiv preprint arXiv:2503.20523}, 2025.

\bibitem[Chang et~al.(2022)Chang, Zhang, Jiang, Liu, and
  Freeman]{chang2022maskgit}
Huiwen Chang, Han Zhang, Lu~Jiang, Ce~Liu, and William~T Freeman.
\newblock Maskgit: Masked generative image transformer.
\newblock In \emph{Proceedings of the IEEE/CVF conference on computer vision
  and pattern recognition}, pages 11315--11325, 2022.

\bibitem[Salimans and Ho(2022)]{salimans2022progressive}
Tim Salimans and Jonathan Ho.
\newblock Progressive distillation for fast sampling of diffusion models.
\newblock \emph{arXiv preprint arXiv:2202.00512}, 2022.

\bibitem[Kodaira et~al.(2025)Kodaira, Hou, Hou, Tomizuka, and
  Zhao]{kodaira2025streamdit}
Akio Kodaira, Tingbo Hou, Ji~Hou, Masayoshi Tomizuka, and Yue Zhao.
\newblock Streamdit: Real-time streaming text-to-video generation.
\newblock \emph{arXiv preprint arXiv:2507.03745}, 2025.

\bibitem[Song et~al.(2023)Song, Dhariwal, Chen, and
  Sutskever]{song2023consistency}
Yang Song, Prafulla Dhariwal, Mark Chen, and Ilya Sutskever.
\newblock Consistency models.
\newblock \emph{International Conference on Machine Learning}, 2023.

\bibitem[Song and Dhariwal(2023)]{song2023improved}
Yang Song and Prafulla Dhariwal.
\newblock Improved techniques for training consistency models.
\newblock \emph{arXiv preprint arXiv:2310.14189}, 2023.

\bibitem[Lu and Song(2024)]{lu2024simplifying}
Cheng Lu and Yang Song.
\newblock Simplifying, stabilizing and scaling continuous-time consistency
  models.
\newblock \emph{arXiv preprint arXiv:2410.11081}, 2024.

\bibitem[Wang et~al.(2023)Wang, Zhang, Zhang, Liu, Zhang, Gao, and
  Sang]{wang2023videolcm}
Xiang Wang, Shiwei Zhang, Han Zhang, Yu~Liu, Yingya Zhang, Changxin Gao, and
  Nong Sang.
\newblock Videolcm: Video latent consistency model.
\newblock \emph{arXiv preprint arXiv:2312.09109}, 2023.

\bibitem[Geng et~al.(2025)Geng, Deng, Bai, Kolter, and He]{geng2025meanflow}
Zhengyang Geng, Mingyang Deng, Xingjian Bai, J~Zico Kolter, and Kaiming He.
\newblock Mean flows for one-step generative modeling.
\newblock \emph{arXiv preprint arXiv:2505.13447}, 2025.

\bibitem[Zhao et~al.(2024)Zhao, Jin, Wang, and You]{zhao2024real}
Xuanlei Zhao, Xiaolong Jin, Kai Wang, and Yang You.
\newblock Real-time video generation with pyramid attention broadcast.
\newblock \emph{arXiv preprint arXiv:2408.12588}, 2024.

\bibitem[Zhang et~al.(2025)Zhang, Chen, Huang, Lin, Liu, Stoica, Xing, and
  Zhang]{zhang2025vsa}
Peiyuan Zhang, Yongqi Chen, Haofeng Huang, Will Lin, Zhengzhong Liu, Ion
  Stoica, Eric Xing, and Hao Zhang.
\newblock Vsa: Faster video diffusion with trainable sparse attention.
\newblock \emph{arXiv preprint arXiv:2505.13389}, 2025.

\bibitem[Damen et~al.(2018)Damen, Doughty, Farinella, Fidler, Furnari, Kazakos,
  Moltisanti, Munro, Perrett, Price, et~al.]{epickitchens}
Dima Damen, Hazel Doughty, Giovanni~Maria Farinella, Sanja Fidler, Antonino
  Furnari, Evangelos Kazakos, Davide Moltisanti, Jonathan Munro, Toby Perrett,
  Will Price, et~al.
\newblock Scaling egocentric vision: The epic-kitchens dataset.
\newblock In \emph{Proceedings of the European conference on computer vision
  (ECCV)}, pages 720--736, 2018.

\end{thebibliography}
\end{hyphenrules}

\clearpage
\appendix

\section{Datasets}

\paragraph{Minecraft VPT}

We use the OpenAI VPT dataset of contractor gameplay \citep{vpt} and combine the available subsets 6--10, resulting in 2541 hours of gameplay.
We split the dataset into 90\% training and 10\% evaluation data, ensuring that the splits do not share any of the same underlying 5-min recording chunks.
We encode keyboard actions as a vector of binary variables and process the mouse actions as in VPT by $\mu$-law encoding, discretizing into 11 bins per coordinate, and enumerating all $11 \times 11 = 121$ combinations to obtain a categorical variable.
The image resolution is $360 \times 640$ and the framerate is 20 FPS.
We zero pad the frames to $384 \times 640$ and then patchify with patch size $16 \times 16$ into 960 tokens.
We reshape the $(N_\mathrm{b} = 512) \times (D_\mathrm{b} = 16)$ bottleneck of the tokenizer to $(N_\mathrm{z} = 256) \times 32$ for the dynamics model.
We train the dynamics model with $N_\mathrm{z} = 256$ spatial tokens, context length $C = 192$, and batch lengths $T_1 = 64$ and $T_2 = 256$.

\paragraph{Minecraft Overworld and Nether split}

To study out-of-distribution generalization of action conditioning in \method, we carefully split the Minecraft dataset into videos of the Overworld versus the Nether dimension.
We also include the End dimension into the Nether portion of the dataset.
Both the Nether and the End feature unique visuals, blocks, and terrain shapes compared to the Overworld.
The Overworld includes natural landscapes with forests, deserts, oceans, and more, whereas the Nether is underworld-themed with red blocks and lava and the End is a space-themed region.
To separate the dataset, we want to ensure no leakage from players entering the Nether/End dimensions and bringing blocks from there back to the Overworld.
For this reason, we exclude the VPT 6 and 7 subsets, which contain long free play.
We then assigned each 5 min recording of the remaining dataset to either the Overworld or the Nether/End portion based on item events that are provided by the dataset.
Whenever a Nether/End item was interacted with, we assign that video to the Nether/End split.
This ensures that the Overworld split contains no Nether/End episodes, whereas the Nether/End split can sometimes contain some Overworld episodes, although this was rare in practice.
We manually investigated the Overworld split obtained by this strategy and found no Nether/End trajectories in it.

\paragraph{SOAR Robotics}

The SOAR dataset \citep{soar} contains teleoperated demonstrations and online trajectories of a reinforcement learning policy, thus covering both successes and failures.
We split the dataset into 90\% training and 10\% evaluation data.
The dataset contains a total of 180 hours of videos with 7D relative end-effector actions.
The image resolution is $256 \times 256$ and the framerate is 5 FPS.
We patchify with patch size $16 \times 16$ into 256 tokens.
We train the dynamics model with $N_\mathrm{z} = 512$ spatial tokens, context length $C = 96$, and batch lengths $T_1 = 32$ and $T_2 = 128$.

\paragraph{Epic Kitchens}

The Epic Kitchens 100 dataset \citep{epickitchens} contains 100 hours of video from the first-person perspective of humans across 45 kitchens.
The test set contains different tasks performed in the same kitchens.
We use the dataset at $256 \times 256$ resolution and 10 FPS.
We patchify with patch size $16 \times 16$ into 256 tokens.
We train the dynamics model with $N_\mathrm{z} = 512$ spatial tokens, context length $C = 96$, and batch lengths $T_1 = 32$ and $T_2 = 128$.

\clearpage
\section{Kitchen Generations}
\vfill
\begin{figure}[h!]
\centering
\includegraphics[width=\linewidth]{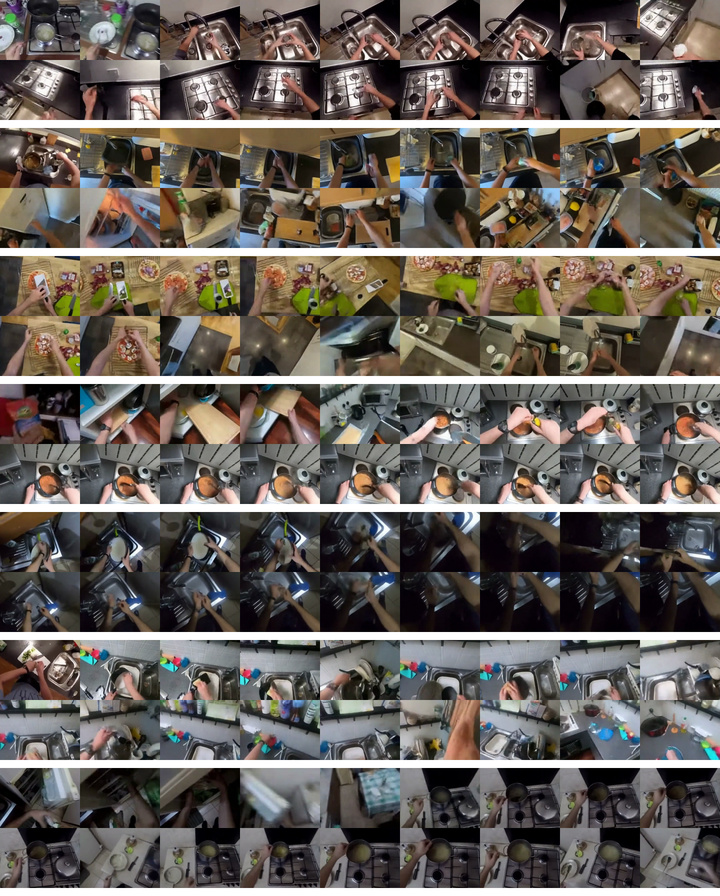}
\caption{Kitchen video generations starting from holdout context.}
\label{fig:kitchens_more}
\end{figure}

\vfill

\section{Minecraft Tasks}
\vspace*{-2ex}
\begin{minipage}[t]{0.48\textwidth}
\centering
\vspace*{1ex}
\begin{mytabular}{
  colspec = {| C{3em} | L{13em} |},
  row{1} = {font=\bfseries},
}

\toprule
Index & Task \\
\midrule
  1 & \texttt{mine\_log} \\
  2 & \texttt{mine\_cobblestone} \\
  3 & \texttt{mine\_iron\_ore} \\
  4 & \texttt{mine\_coal} \\
  5 & \texttt{mine\_diamond} \\
  6 & \texttt{craft\_planks} \\
  7 & \texttt{craft\_stick} \\
  8 & \texttt{craft\_crafting\_table} \\
  9 & \texttt{craft\_furnace} \\
 10 & \texttt{craft\_iron\_ingot} \\
 11 & \texttt{craft\_wooden\_pickaxe} \\
 12 & \texttt{craft\_stone\_pickaxe} \\
 13 & \texttt{craft\_iron\_pickaxe} \\
 14 & \texttt{open\_crafting\_table} \\
 15 & \texttt{open\_furnace} \\
 16 & \texttt{place\_crafting\_table} \\
 17 & \texttt{place\_furnace} \\
 18 & \texttt{use\_wooden\_pickaxe} \\
 19 & \texttt{use\_stone\_pickaxe} \\
 20 & \texttt{use\_iron\_pickaxe} \\
\bottomrule

\end{mytabular}
\captionof{table}{Taskset for training the multi-task agent.}
\label{tab:mc_tasks}

\vspace*{3ex}
\centering
\newcommand{\icon}[1]{\includegraphics[height=2.5ex]{icons/#1.png}}
\newcommand{\name}[1]{\raisebox{.4ex}{#1}}
\vspace*{1ex}
\begin{mytabular}{
  colspec = {| C{3em} | L{13em} |},
  row{1} = {font=\bfseries},
}

\toprule
Icon & Item \\
\midrule
\icon{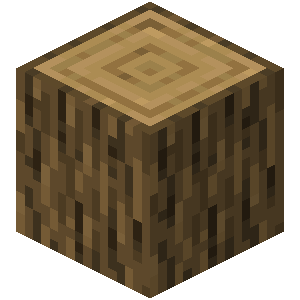} & \name{Log} \\
\icon{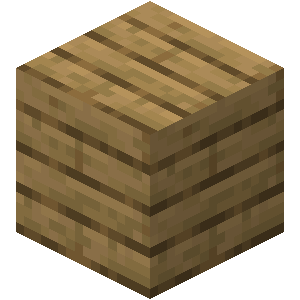} & \name{Planks} \\
\icon{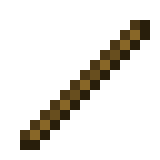} & \name{Stick} \\
\icon{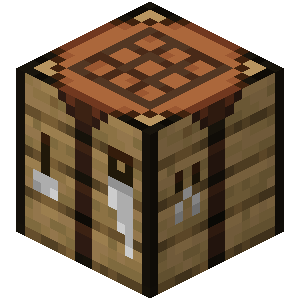} & \name{Crafting table} \\
\icon{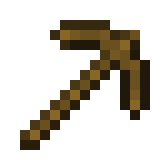} & \name{Wooden pickaxe} \\
\icon{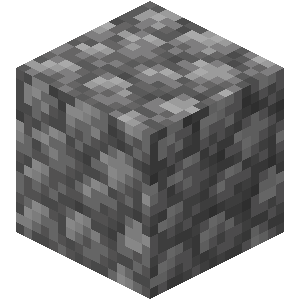} & \name{Cobblestone} \\
\icon{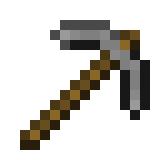} & \name{Stone pickaxe} \\
\icon{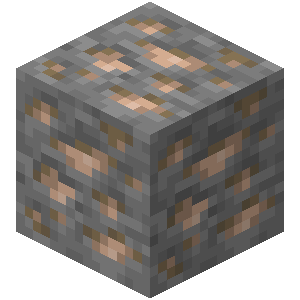} & \name{Iron ore} \\
\icon{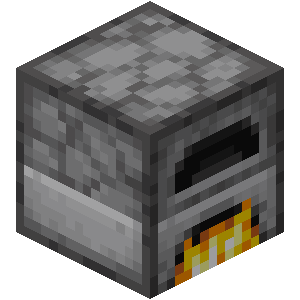} & \name{Furnace} \\
\icon{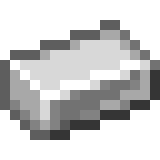} & \name{Iron ingot} \\
\icon{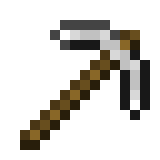} & \name{Iron pickaxe} \\
\icon{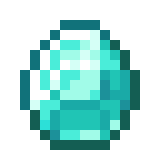} & \name{Diamond} \\
\bottomrule

\end{mytabular}
\captionof{table}{Milestone items used for measuring progress during evaluation.}
\label{tab:mc_items}

\end{minipage}
\hfill%
\begin{minipage}[t]{0.48\textwidth}
\centering
\vspace*{1ex}
\begin{mytabular}{
  colspec = {| L{13em} | C{3em} |},
  row{1} = {font=\bfseries},
}

\toprule
Task & Count \\
\midrule

\texttt{mine\_log} & 10 \\
\texttt{craft\_planks} & 20 \\
\texttt{craft\_crafting\_table} & 1 \\
\texttt{place\_crafting\_table} & 1 \\
\texttt{craft\_stick} & 4 \\
\texttt{craft\_wooden\_pickaxe} & 1 \\
\texttt{use\_wooden\_pickaxe} & 1 \\
\texttt{mine\_cobblestone} & 3 \\
\texttt{craft\_planks} & 4 \\
\texttt{craft\_crafting\_table} & 1 \\
\texttt{place\_crafting\_table} & 1 \\
\texttt{craft\_stick} & 4 \\
\texttt{craft\_stone\_pickaxe} & 1 \\
\texttt{use\_stone\_pickaxe} & 1 \\
\texttt{mine\_iron\_ore} & 3 \\
\texttt{mine\_cobblestone} & 8 \\
\texttt{craft\_planks} & 4 \\
\texttt{craft\_crafting\_table} & 1 \\
\texttt{place\_crafting\_table} & 1 \\
\texttt{craft\_furnace} & 1 \\
\texttt{craft\_iron\_ingot} & 3 \\
\texttt{craft\_stick} & 2 \\
\texttt{craft\_iron\_pickaxe} & 1 \\
\texttt{use\_iron\_pickaxe} & 1 \\
\texttt{mine\_diamond} & $\infty$ \\

\bottomrule

\end{mytabular}
\captionof{table}{Prompt sequence for evaluation.}
\label{tab:mc_ladder}

\end{minipage}
\clearpage

\section{Offline Diamond Challenge}
\begin{table}[h!]
\centering
\newcommand{\rot}[1]{\hspace*{1ex}\rotatebox{70}{#1}}
\begin{mytabular}{
  colspec = {| L{8em} | C{3em} C{3em} C{3em} C{3em} C{3em} C{3em} C{3em} C{3em} |},
  row{1} = {font=\bfseries},
  stretch=0.9,
}
\toprule
Item & \rot{VPT (pretrained)} & \rot{VPT (finetuned)} & \rot{BC (notask)} & \rot{WM+BC (notask)} & \rot{BC} & \rot{VLA (Gemma 3)} & \rot{WM+BC} & \rot{Dreamer 4} \\
\midrule
Log            &           \o81.9  &           \o84.3  &           \o71.4  &           \o92.6  &   \textbf{\o97.3} &   \textbf{\o98.5} &   \textbf{\o99.6} &   \textbf{\o99.1} \\
Planks         &           \o30.6  &           \o65.3  &           \o68.6  &           \o91.6  &   \textbf{\o95.7} &   \textbf{\o98.3} &   \textbf{\o99.6} &   \textbf{\o98.9} \\
Crafting table &          \o\o1.7  &          \o\o4.7  &           \o63.8  &           \o90.6  &           \o93.5  &   \textbf{\o97.2} &   \textbf{\o99.1} &   \textbf{\o98.5} \\
Stick          &           \o30.3  &           \o52.6  &           \o62.4  &           \o90.1  &   \textbf{\o95.0} &   \textbf{\o97.7} &   \textbf{\o98.9} &   \textbf{\o98.7} \\
Wooden pickaxe &          \o\o0.0  &          \o\o0.0  &           \o33.8  &           \o77.3  &           \o86.5  &   \textbf{\o94.1} &   \textbf{\o97.3} &   \textbf{\o96.6} \\
Cobblestone    &          \o\o4.8  &          \o\o6.9  &           \o32.0  &           \o77.4  &           \o83.9  &           \o91.6  &   \textbf{\o97.2} &   \textbf{\o95.9} \\
Stone pickaxe  &          \o\o0.0  &          \o\o0.0  &          \o\o8.8  &           \o38.4  &           \o53.8  &           \o76.7  &   \textbf{\o89.4} &   \textbf{\o90.1} \\
Iron ore       &          \o\o0.1  &          \o\o0.1  &          \o\o3.6  &           \o22.0  &           \o26.5  &           \o46.3  &           \o62.9  &   \textbf{\o66.7} \\
Furnace        &          \o\o0.0  &          \o\o0.0  &          \o\o4.0  &           \o28.0  &           \o16.2  &           \o42.4  &           \o51.1  &   \textbf{\o58.1} \\
Iron ingot     &          \o\o0.1  &          \o\o0.1  &          \o\o0.2  &          \o\o1.2  &          \o\o4.3  &           \o22.5  &           \o27.8  &   \textbf{\o39.5} \\
Iron pickaxe   &          \o\o0.0  &          \o\o0.0  &          \o\o0.0  &          \o\o0.1  &          \o\o0.6  &           \o11.2  &           \o16.9  &   \textbf{\o29.0} \\
Diamond        &          \o\o0.0  &          \o\o0.0  &          \o\o0.0  &          \o\o0.0  &          \o\o0.0  &          \o\o0.0  &          \o\o0.0  &  \textbf{\o\o0.7} \\
\bottomrule
\end{mytabular}
\caption{Success rates for each milestone item averaged over 1000 evaluation episodes. Scores within 5\% of the highest recorded score are highlighted in bold.}
\label{tab:rl_success}
\end{table}

\begin{table}[h!]
\centering
\newcommand{\rot}[1]{\hspace*{1ex}\rotatebox{70}{#1}}
\begin{mytabular}{
  colspec = {| L{8em} | C{3em} C{3em} C{3em} C{3em} C{3em} C{3em} C{3em} C{3em} |},
  row{1} = {font=\bfseries},
  stretch=0.9,
}
\toprule
Item & \rot{VPT (pretrained)} & \rot{VPT (finetuned)} & \rot{BC (notask)} & \rot{WM+BC (notask)} & \rot{BC} & \rot{VLA (Gemma 3)} & \rot{WM+BC} & \rot{Dreamer 4} \\
\midrule
Log            &          \o\o9.1  &          \o\o6.3  &           \o11.9  &          \o\o5.4  &          \o\o1.8  &          \o\o2.2  &          \o\o1.2  &  \textbf{\o\o0.9} \\
Planks         &           \o25.2  &           \o14.2  &           \o12.2  &          \o\o5.9  &          \o\o4.3  &          \o\o3.4  &          \o\o2.1  &  \textbf{\o\o2.0} \\
Stick          &           \o32.0  &           \o24.0  &           \o13.3  &          \o\o6.7  &          \o\o6.4  &          \o\o5.0  &  \textbf{\o\o3.1} &  \textbf{\o\o2.9} \\
Crafting table &           \o41.4  &           \o27.5  &           \o17.1  &          \o\o8.0  &          \o\o9.5  &          \o\o7.2  &  \textbf{\o\o4.6} &  \textbf{\o\o4.4} \\
Wooden pickaxe &               --- &               --- &           \o18.8  &           \o11.6  &           \o11.4  &          \o\o9.8  &          \o\o5.7  &  \textbf{\o\o5.0} \\
Cobblestone    &               --- &               --- &           \o19.6  &           \o12.7  &           \o13.3  &           \o12.1  &          \o\o6.7  &  \textbf{\o\o5.6} \\
Stone pickaxe  &               --- &               --- &           \o23.5  &           \o15.7  &           \o15.8  &           \o14.5  &          \o\o8.9  &  \textbf{\o\o6.7} \\
Iron ore       &               --- &               --- &           \o28.9  &           \o17.5  &           \o20.9  &           \o23.5  &           \o14.3  &  \textbf{\o\o9.9} \\
Furnace        &               --- &               --- &           \o29.4  &           \o19.7  &           \o24.5  &           \o24.7  &           \o16.1  &   \textbf{\o11.0} \\
Iron ingot     &               --- &               --- &               --- &           \o30.5  &           \o28.8  &           \o30.8  &           \o17.2  &   \textbf{\o12.4} \\
Iron pickaxe   &               --- &               --- &               --- &               --- &           \o29.1  &           \o31.1  &           \o17.0  &   \textbf{\o13.3} \\
Diamond        &               --- &               --- &               --- &               --- &               --- &               --- &               --- &   \textbf{\o20.7} \\
\bottomrule
\end{mytabular}
\caption{Time in minutes needed for reaching each milestone item, averaged over successful episodes. We omit timings for items below a success rate of 0.5\% to ensure statistical significance. Scores within 5\% of the fastest recorded score are highlighted in bold.}
\label{tab:rl_timing}
\end{table}

\clearpage

\enlargethispage{\baselineskip}
\vspace*{-9ex}
\section{Minecraft Inputs}
\vspace*{-1ex}
\begin{figure}[h!]
\centering
\begin{subfigure}[b]{0.25\textwidth}
\includegraphics[width=\textwidth]{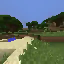}
\caption{Dreamer 3 (64$\times$64)}
\end{subfigure}\hfill%
\begin{subfigure}[b]{0.25\textwidth}
\includegraphics[width=\textwidth]{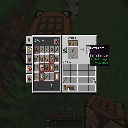}
\caption{VPT (128$\times$128)}
\end{subfigure}\hfill%
\begin{subfigure}[b]{0.445\textwidth}
\includegraphics[width=\textwidth]{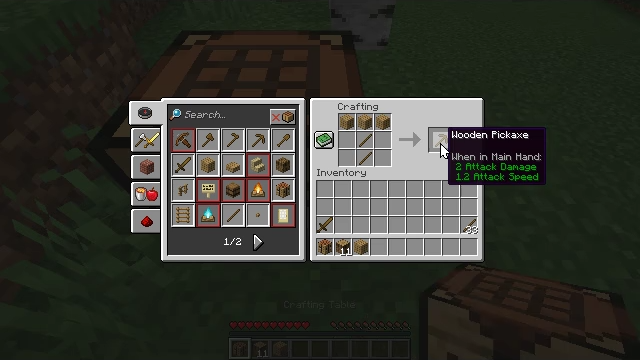}
\caption{\method (360$\times$640)}
\end{subfigure}%
\caption{%
Comparison of input images for different agents.
\method learns directly from high-resolution images reflecting the experience of human players.
}
\label{fig:inputs}
\end{figure}

\section{Previous Dreamer Generations}
\vspace*{-1ex}
Dreamer~3\citep{dreamerv3} learned to obtain diamonds in Minecraft from scratch by online interaction.
Its inputs are low-resolution images and inventory states and the outputs are mouse, keyboard, and abstract crafting actions.
Dreamer~3 uses a recurrent state-space model (RSSM) \citep{hafner2018planet} as its world model, which is based on a recurrent neural network and a variational objective.
This approach results in a lightweight world model with highly efficient inference but is difficult to scale to diverse data distributions.
In contrast, \method learns to obtain diamonds in Minecraft purely from offline data.
Its inputs are only high-resolution images and the outputs are low-level mouse and keyboard actions.
\method uses a scalable world model based on an efficient transformer architecture and a shortcut forcing objective, allowing it to scale to diverse data distributions with many details.
While Dreamer~3 uses return normalization and an entropy regularizer, \method uses PMPO with a KL to the behavioral cloning prior for imagination training, where no normalization is needed.
\vspace*{1ex}
\begin{figure}[h!]
\centering
\includegraphics[width=\linewidth]{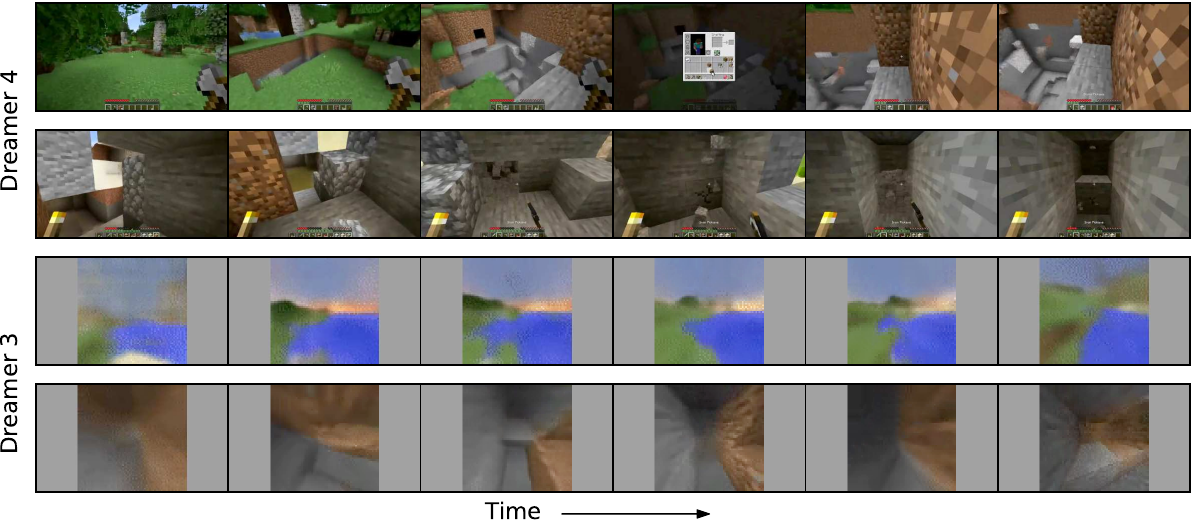}
\vspace*{-3ex}
\caption{Comparison of multi-step video generations between Dreamer~3 and \method.
}
\label{fig:openl}
\vspace*{-1ex}
\end{figure}

\clearpage

\enlargethispage{\baselineskip}
\vspace*{-10ex}
\section{Human Interaction: Lucid-v1}
\vspace*{-2ex}
\begin{figure}[h!]
\includegraphics[width=0.97\linewidth]{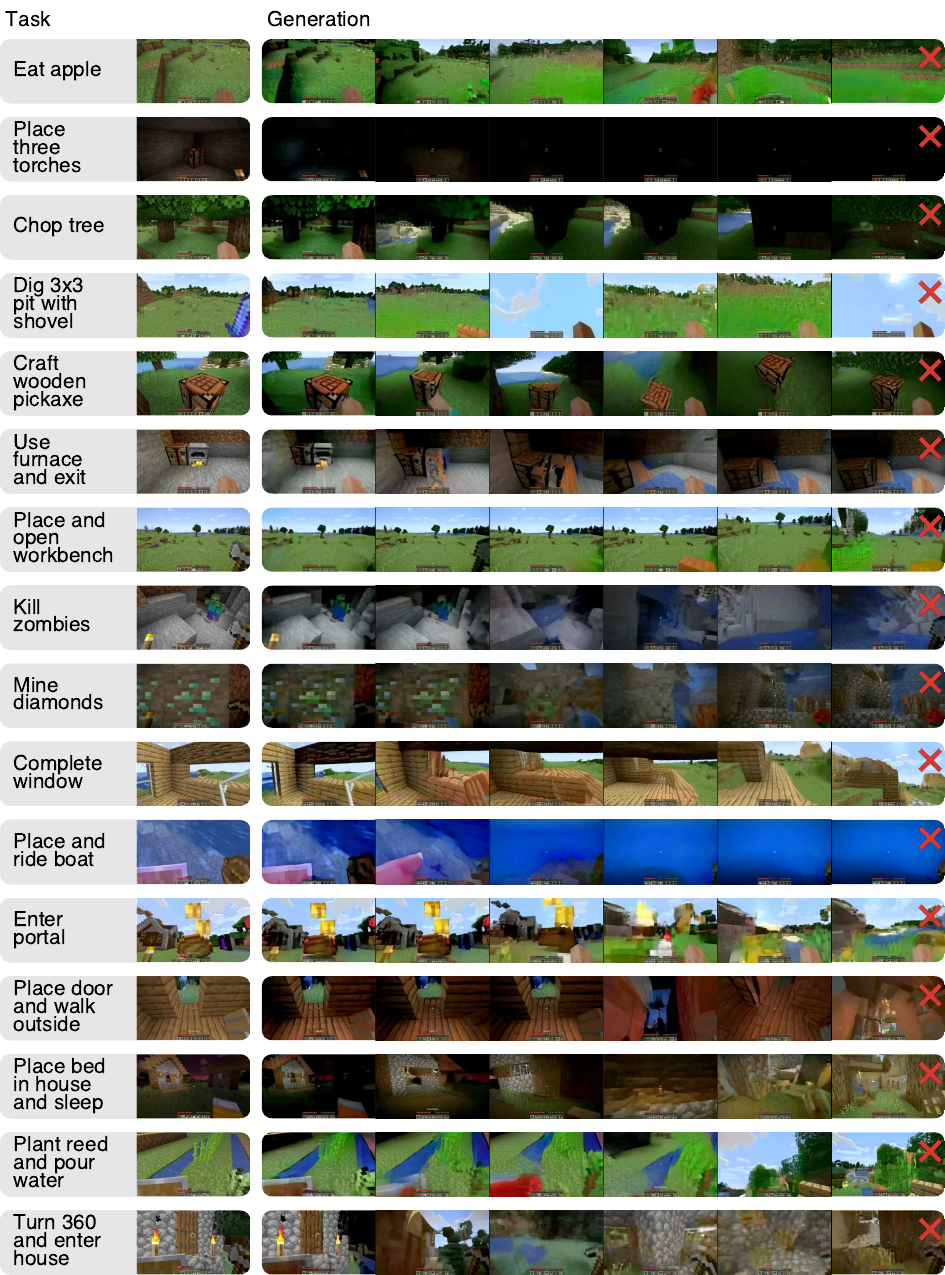}
\caption{Lucid-v1}
\label{fig:wmtasks_lucid}
\end{figure}
\vspace*{-2ex}
\clearpage

\enlargethispage{\baselineskip}
\vspace*{-10ex}
\section{Human Interaction: OASIS (large)}
\vspace*{-2ex}
\begin{figure}[h!]
\includegraphics[width=0.97\linewidth]{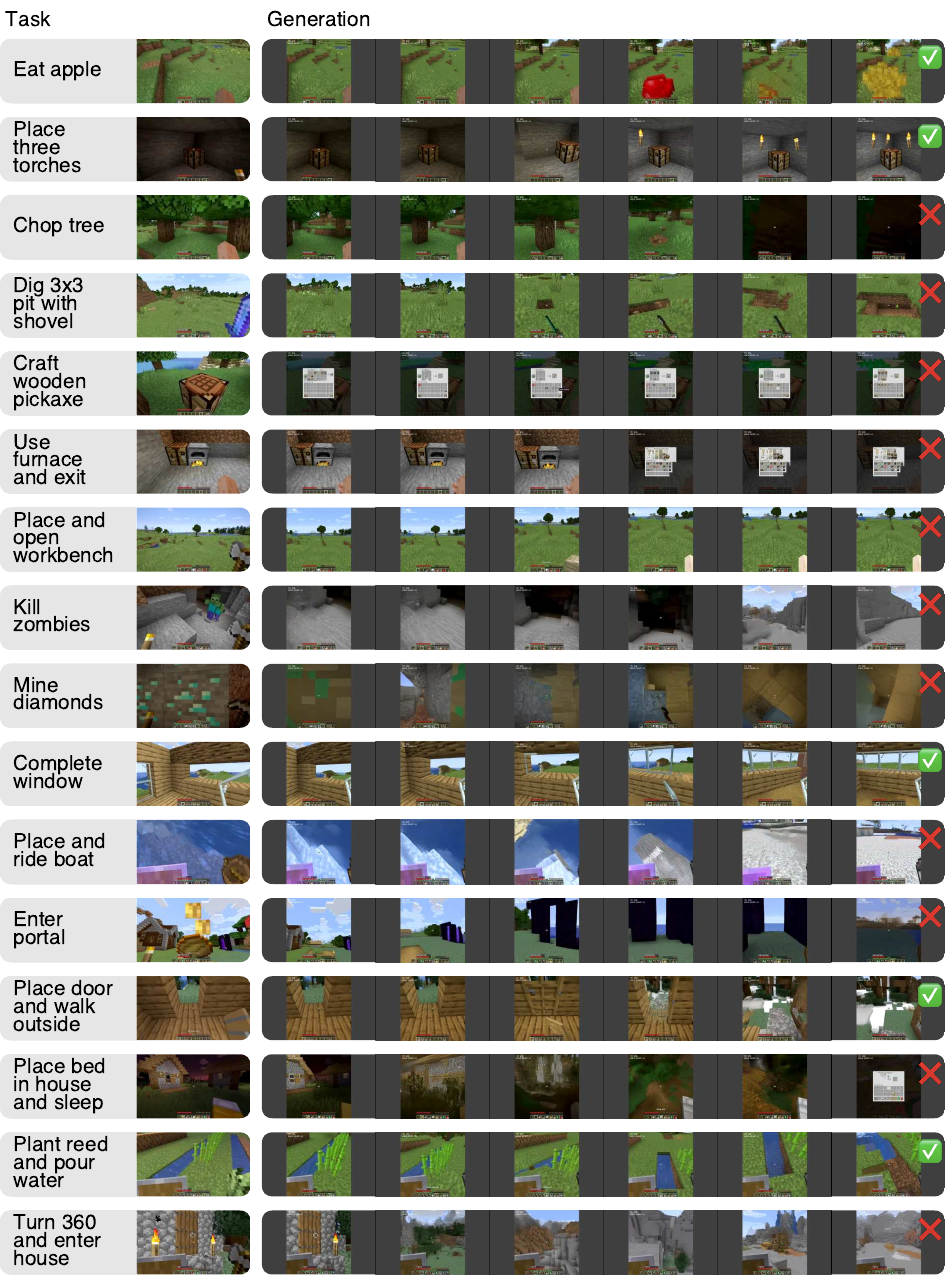}
\caption{Oasis (large)}
\label{fig:wmtasks_oasis}
\end{figure}
\vspace*{-2ex}
\clearpage

\enlargethispage{\baselineskip}
\vspace*{-10ex}
\section{Human Interaction: \method}
\vspace*{-2ex}
\begin{figure}[h!]
\includegraphics[width=0.97\linewidth]{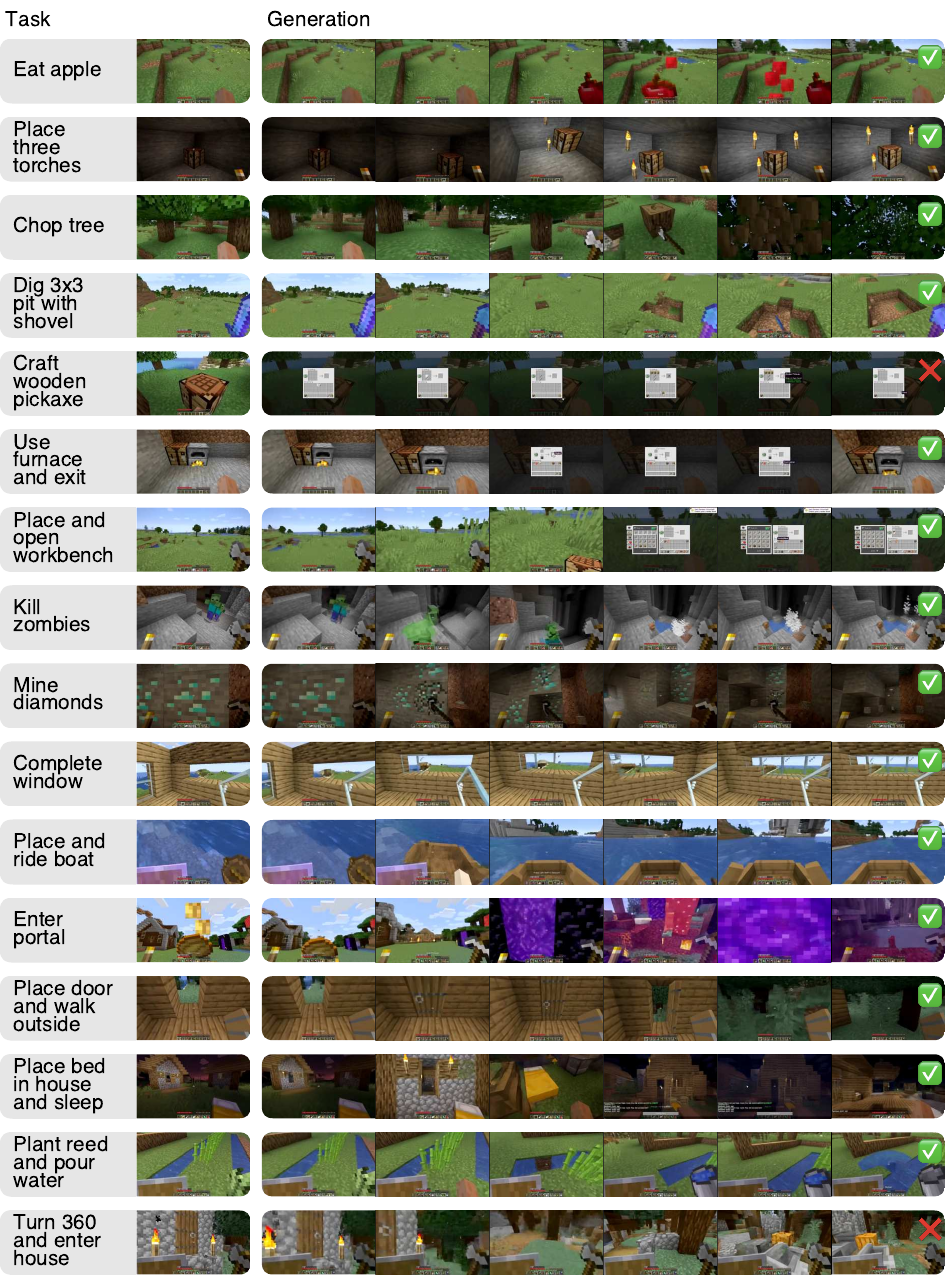}
\caption{\method}
\label{fig:wmtasks_ours}
\end{figure}
\clearpage

\end{document}